\documentclass[twoside,11pt]{article}

\usepackage{jmlr2e}
\usepackage[utf8]{inputenc}
\usepackage{amsmath}
\usepackage{url}
\usepackage{color}
\usepackage{enumitem}
\usepackage{widetable}
\usepackage{booktabs}
\usepackage{epigraph}
\usepackage{setspace}
\usepackage{multirow}
\usepackage[ruled]{algorithm2e} 
\usepackage{xcolor} 

\makeatletter
\renewcommand\paragraph{\@startsection{paragraph}{4}{\z@}%
                                    {1.5ex \@plus1ex \@minus.2ex}%
                                    {-1em}%
                                    {\normalfont\normalsize\itshape}}
\makeatother

\firstpageno{1}

\title{Compute-Efficient Deep Learning: Algorithmic Trends and Opportunities}

\usepackage{lastpage}
\jmlrheading{24}{2023}{1-\pageref{LastPage}}{10/22; Revised
2/23}{3/23}{22-1208}{Brian R. Bartoldson, Bhavya Kailkhura, Davis Blalock}
\ShortHeadings{Compute-Efficient Deep Learning}{Bartoldson, Kailkhura, and Blalock}
\begin{document}

\author{\name Brian R. Bartoldson\thanks{All authors contributed equally to this research. Bhavya led the study conceptualization and taxonomy design. Brian led the written survey of the literature. Davis conducted all experiments and led the creation of a guide to achieving speedups in practice.}
\email Bartoldson@llnl.gov \\
\addr Lawrence Livermore National Laboratory, USA
\AND
\name Bhavya Kailkhura$^\ast$
\email Kailkhura1@llnl.gov \\
\addr Lawrence Livermore National Laboratory, USA
\AND
\name Davis Blalock$^\ast$
\email Davis@MosaicML.com \\
\addr MosaicML, USA
}

\editor{Samy Bengio}

\maketitle

\begin{abstract}%
Although deep learning has made great progress in recent years, the exploding economic and environmental costs of training neural networks are becoming unsustainable. 
To address this problem, there has been a great deal of research on \emph{algorithmically-efficient deep learning}, which seeks to reduce training costs not at the hardware or implementation level, but through changes in the semantics of the training program. 
In this paper, we present a structured and comprehensive overview of the research in this field.
First, we formalize the \emph{algorithmic speedup} problem, then we use fundamental building blocks of algorithmically efficient training to develop a taxonomy. 
Our taxonomy highlights commonalities of seemingly disparate methods and reveals current research gaps. Next, we present evaluation best practices to enable comprehensive, fair, and reliable comparisons of speedup techniques. To further aid research and applications, we discuss common bottlenecks in the training pipeline (illustrated via experiments) and offer taxonomic mitigation strategies for them. Finally, we highlight some unsolved research challenges and present promising future directions.
\end{abstract}

\begin{keywords}
deep learning, training speedup, computational efficiency, carbon emission
\end{keywords}

\section{Introduction}

\setlength{\epigraphwidth}{0.55\textwidth}
\epigraph{``\textit{Science is a way of thinking much more than it is a body of knowledge.}''}{--- Carl Sagan}

In the last few years, deep learning (DL) has made significant progress on a wide range of applications, such as protein structure prediction (AlphaFold, \citeauthor{2021alphafold}, \citeyear{2021alphafold}), text-to-image synthesis (DALL-E, \citeauthor{2021dalle}, \citeyear{2021dalle}), text generation (GPT-3, \citeauthor{2022gpt3}, \citeyear{2022gpt3}), etc. 
The key strategy behind achieving these performance gains is scaling up DL models to extremely large sizes and training them on massive amounts of data. 
For most applications, the number of trainable parameters is doubling at least every 18 to 24 months---language models are leading with a 4- to 8-month doubling time \citep{parametercount}. Notable examples of massive AI models include the following: Swin Transformer-V2 \citep{2022swin} with 3 billion parameters for vision applications, PaLM \citep{2022palm} with 540 billion parameters for language modeling, and Persia \citep{2021persia} with 100 trillion parameters for content recommendations.

Although scaling up DL models is enabling unprecedented advances, training large models has become extremely expensive. For example, GPT-3 training was estimated to cost \$1.65 million with Google v3 TPUs \citep{powerai} and inefficient/naive development of a transformer model would emit carbon dioxide (CO2) equivalent to the lifetime carbon footprint of five cars \citep{strubell2019energy}. Concerningly, DL has still not reached the performance level required by many of its applications: e.g., human-level performance is required for deploying fully autonomous vehicles in the real world but hasn't yet been reached.
Growing model and data sizes to reach such required performances will make current training strategies unsustainable on financial, environmental, and other fronts. Indeed, extrapolating current trends, the training cost of the largest AI model in 2026 would be more than the total U.S. GDP \citep{powerai}.
Moreover, the heavy compute reliance of DL raises concerns around the marginalization of users with limited financial resources like academics, students, and researchers (particularly those from emerging economies) \citep{ahmed2020democratization}. We discuss these critical issues in more detail in Appendix~\ref{app:motivation}.

Given the unsustainable growth of its computational burden, progress with DL demands more compute-efficient training methods. A natural direction is to eliminate algorithmic inefficiencies in the learning process to reduce the time, cost, energy, and carbon footprint of DL training. Such \emph{Algorithmically-Efficient Deep Learning} methods could change the training process in a variety of ways that include: altering the data or the order in which samples are presented to the model; tweaking the structure of the model; and changing the optimization algorithm. These algorithmic improvements are critical to moving towards estimated lower bounds on the required computational burden of effective DL training, which are greatly exceeded by the burden induced by current practices \citep{thompson2020computational}. Further, these algorithmic gains compound with software and hardware acceleration techniques \citep{hernandez2020measuring}. Thus, we believe algorithmically-efficient DL presents an enormous opportunity to increase the benefits of DL and reduce its costs.

While this view is supported by the recent surge in algorithmic efficiency papers, these papers also suggest that research and application of algorithmic efficiency methods are hindered by fragmentation. 
Disparate metrics are used to quantify efficiency, which produces inconsistent rankings of speedup methods. Evaluations are performed on narrow or poorly characterized environments, which results in incorrect  or overly-broad conclusions. Algorithmic efficiency methods are discussed in the absence of a taxonomy that reflects their breadth and relationships, which makes it hard to understand how to traverse the speedup landscape to combine different methods and develop new ones.

To address these fragmentation issues, we eschew a more traditional survey approach that focuses on just a single component (e.g., the model) or single action (e.g., reducing model size) in the training pipeline. Instead, we adopt a wholistic view of the speedup problem and emphasize that one needs to carefully select a combination of techniques, which we survey in Section 3, to overcome various compute-platform bottlenecks. We use experiments to illustrate the importance of such a wholistic view to achieving speedup in practice, and we provide guidance informed by the relationships between different bottlenecks and components of training. While this leads to our consideration of a broad set of topics, we limit our scope to training (not inference), algorithms (not efficient hardware or compute-kernels), trends (not every possible method), and techniques applicable to language and computer vision tasks (rather than techniques specific to RL, graphs, etc.).

Accordingly, our central contributions are an organization of the algorithmic-efficiency literature (via a taxonomy and survey inspired by \citeauthor{von2019informed}, \citeyear{von2019informed}) and technical characterization of the practical issues affecting the reporting and achievement of speedups (via guides for evaluation and practice). Throughout, our discussion emphasizes the critical intersection of these two thrusts: e.g., whether an algorithmic efficiency method leads to an actual speedup indeed depends on the interaction of the method (understandable via our taxonomy) and the compute platform (understandable via our practitioner's guide). Our contributions are summarized as follows:

\begin{itemize}
    \item \textbf{Formalizing Speedup:} We review DNN efficiency metrics, then formalize the algorithmic speedup problem.
    \item \textbf{Taxonomy and Survey:} We classify over 200 papers via 5 speedup actions (the 5Rs) that apply to 3 training-pipeline components (see Tables \ref{tab:main_approaches} and \ref{tab:class}).
    The taxonomy facilitates selection of methods for practitioners, digestion of the literature for readers, and identification of opportunities for researchers.
    \item \textbf{Best Evaluation Practices:} We identify evaluation pitfalls common in the literature and accordingly present best evaluation practices to enable comprehensive, fair, and reliable comparisons of various speedup techniques.
    \item \textbf{Practitioner's Guide:} We discuss compute-platform bottlenecks that affect speedup-method effectiveness. Connecting our survey and practice, we suggest appropriate methods and mitigations based on the location of the bottlenecks in the training pipeline.
\end{itemize}


\begin{table}
\centering
{\setstretch{0.75} \fontsize{7.5pt}{10.25pt}\selectfont{
\begin{tabular}{p{.1\textwidth}p{.1\textwidth}p{.05\textwidth}p{.15\textwidth}p{.18\textwidth}p{.26\textwidth}}
    \toprule
    \multicolumn{4}{c}{\centering {\small \textbf{Popular Speedup Approaches}}}
    & 
    \multicolumn{2}{c}{\centering {\small \textbf{Example Details}}}
    \\
    \cmidrule(lr){1-4}
    \cmidrule(lr){5-6}
    Component &
    Subcomponent &
    Action &
    Example &
    Key Challenge &
    Future Direction
    \\
    \midrule
    \multirow{6.5}{*}{\hfil Function }
    &
    \multirow{3}{*}{\hfil \shortstack[l]{Model \\ parameters  }}
    &
    \multirow{3}{*}{\hfil Restrict }
    &
    Structured parameterization (e.g., factorization)
    &
    Identifying hardware-efficient patterns
    &
    Dynamic approaches, Hardware-aware learnable parameterization
    \\
    \cmidrule{2-6}
    &
    \multirow{3}{*}{\hfil Architecture }
    &
    \multirow{3}{*}{\hfil Remove }
    &
    Remove expensive layers (e.g., NFNets)
    &
    Maintaining the performance and stability of the original model
    &
    Learning-based and dynamic approaches to identify architectural redundancies 
    \\
    \midrule

    \multirow{9.5}{*}{\hfil Data }
    &
    \multirow{4}{*}{\hfil \shortstack[l]{Training \\    Data  }}
    &
    \multirow{4}{*}{\hfil Reorder }
    &
    Curriculum learning (e.g., teaching with commentaries)
    &
    Automatic design of curriculum for new tasks/problems
    &
    Meta-learning curriculum for larger-scale datasets and new applications
    \\
    \cmidrule{2-6}
    &
    \multirow{5}{*}{\hfil \shortstack[l]{    Derived  \\    Data  }}
    &
    \multirow{5}{*}{\hfil Remove }
    &
    Gradient pruning (e.g., Selective Backprop)
    &
    Maintaining generalization performance on challenging datasets, e.g., ImageNet
    &
    Accurate selection strategies for complex datasets (e.g., ImageNet) and challenging problems (e.g., object detection)
    \\
    \midrule

    \multirow{6.5}{*}{\hfil Optimization }
    &
    \multirow{3}{*}{\hfil \shortstack[l]{Training \\    Objective  }}
    &
    \multirow{3}{*}{\hfil Retrofit }
    &
    Adding curvature information (e.g., SAM)
    &
    Reducing the costliness of computing new update
    &
    Function approximation for reducing cost while maintaining the accuracy of new updates 
    \\
    \cmidrule{2-6}
    &
    \multirow{3}{*}{\hfil \shortstack[l]{Training \\    Algorithm  }}
    &
    \multirow{3}{*}{\hfil Replace }
    &
    Using a better optimizer (e.g., learned optimizer)
    &
    Reducing instability and computational inefficiency  
    &
    Scalable and stable training methods for bigger and more complicated models
    \\
    \bottomrule
\end{tabular}
}}
\caption{
    \textbf{Popular Approaches to Speedup.} 
    Speedup techniques can be categorized in terms of ``5R'' actions applicable to three components of the training pipeline. 
    For each component of the DNN training pipeline, we report an example speedup approach found in our literature review and corresponding details.
}
\label{tab:main_approaches}
\end{table}

With these contributions, we hope to improve the research and application of algorithmic efficiency, a critical piece of the compute-efficient deep learning needed to overcome the economic, environmental, and inclusion-related roadblocks faced by existing research. This paper is organized mainly into four parts: 
Section \ref{sec:definition} provides an overview of DNN training and efficiency metrics along with a formalization of the algorithmic speedup problem. Section \ref{sec:taxonomy} uses broadly applicable building blocks of speedup methods and the training pipeline components they affect to develop our taxonomy.   
Section \ref{sec:survey} presents a comprehensive categorization of the speedup literature based on our taxonomy and discusses research opportunities and challenges.
Sections \ref{sec:evaluation} and \ref{sec:guide} discuss best evaluation practices for comparing different approaches and our practical recommendations for choosing suitable speedup methods, respectively.
Finally, Section \ref{sec:conclusion} concludes and presents open questions in the algorithmic-efficiency area.

\section{Compute-Efficient Training: Overview, Metrics, and Definition}
\label{sec:definition}
In this section, we first provide a brief overview of the Deep Neural Network (DNN) training process. Next, we mention various metrics that quantify training efficiency and discuss their pros and cons. Finally, we formally define algorithmic speedup for DNN training.

\subsection{Overview of DNN Training Process}
\begin{figure}[t]
\centering
\includegraphics[width=1\linewidth]{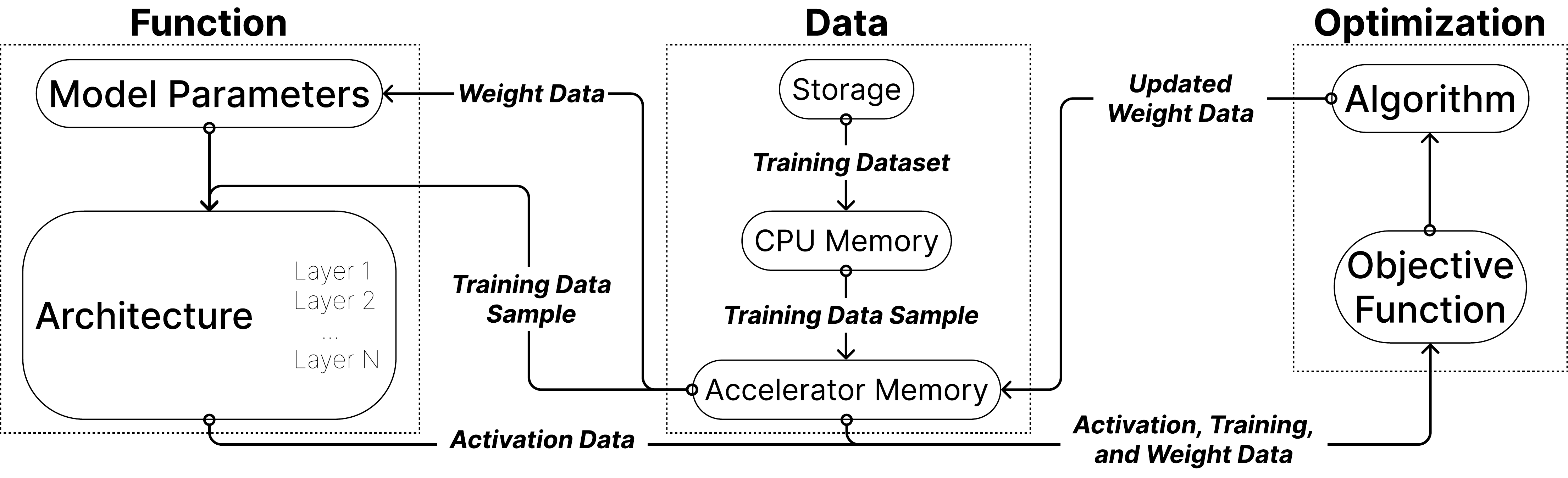}
\caption{Illustration of data movement and computational steps needed to complete a typical iteration of neural network training.}
\label{fig:pipeline}
\end{figure}

At a high level, the goal of DL is to learn a function that can map inputs to outputs to accomplish a certain task. 
This function, referred to as the model, is chosen from a parametric family called the \textit{architecture} during training.
Model training is formulated as an optimization problem, where the goal of the optimizer is to find model parameters that optimize the training objective function.  On each iteration, a DNN training algorithm typically performs the following steps (see Figure~\ref{fig:pipeline}):
\begin{itemize}
    \item Load a subset of the training dataset (referred to as a batch) from the storage device into the main memory. Apply necessary preprocessing on the CPU---e.g., JPEG decompression, data transformation, and data augmentation.
    \item Move the preprocessed and batched data samples to the AI accelerator (e.g., a GPU) to perform forward and backward propagation, thereby obtaining derived data such as activations and gradients:
    \begin{itemize}
        \item \textit{Forward Pass:} traverse the computational graph in the direction of dependencies defined by the DNN. To enable the backward pass, intermediate outputs (i.e., activations) preceding trainable or nonlinear operations must be stored.
        \item \textit{Backward Pass:} traverse the computational graph in reverse order, calculating (using the chain rule) and storing the gradients for intermediate variables and parameters.
    \end{itemize}
    \item Compute the descent direction using the gradients from the backpropagation step and any state variables kept by the optimization algorithm. Combine this direction with the learning rate (step size) to update the model parameters and adjust the optimizer state as needed.
\end{itemize}

This iteration is repeated until the desired model quality or a predefined number of iterations is reached.
Thus, total training time $T_{\text{total}}$ can be calculated as:

\begin{equation*}
    T_{\text{total}} =  \sum_{i=1}^{n_{\text{iteration}}} T_i,
    \end{equation*}
where $T_i$ is the time taken to finish iteration $i$ and $n_{\text{iteration}}$ is the total number of training iterations required. 

Recent increases in $T_{\text{total}}$ within modern training pipelines are outpacing advancements in compute platforms, accenting the need for algorithmically-efficient solutions \citep{patterson2022carbon,menghani2021efficient}. Devising such efficient training procedures---e.g., to minimize redundant computations in areas that are bottlenecks on the compute platform ---is thus a critical direction to explore. 

\subsection{Metrics to Quantify Training Efficiency}
In order to devise efficient training methods, we first need a proper measure of efficiency. In this section, we briefly summarize some commonly used efficiency metrics along with their benefits and drawbacks. For a more detailed discussion on this topic, we refer readers to \cite{schwartz2020green} and \cite{dehghani2021efficiency}.

\paragraph{Training Time.} A direct measure for evaluating efficiency is training time. This metric has the advantage of being simple to measure and coupled to real-world utility. It is also proportional to monetary cost when training on rented hardware and related to hardware depreciation when training on hardware one owns. The downside of training time is that it depends on one's precise hardware and software configuration, which makes comparison across papers dubious at best.

\paragraph{FLOPs.} Another popular metric is the total number of floating-point operations (FLOPs) involved in a model's forward pass or the entire training process. FLOPs do not have a formal, universal definition. However, they typically reflect the number of elementary arithmetic operations present in a computation, including multiplications, additions, divisions, subtractions, and transcendental functions.
An appealing property of FLOPs is that they can be computed without taking into account the hardware. This allows comparison across different hardware and software stacks. Unfortunately, FLOPs fail to capture factors such as memory access time, communication overhead, the instructions available on particular hardware, compute utilization, and more.

\paragraph{Number of Model Parameters.} Another common measure of computational cost is the number of model parameters. Similar to FLOPs, this metric is agnostic to the hardware on which the model is being trained. Moreover, it also determines the amount of memory consumed by the model and optimizer state. However, like FLOPs, it does not account for most factors affecting cost and runtime. Further, different models with similar parameter counts may need different amounts of work to achieve a certain performance level; e.g., increasing the input's resolution or sequence length increases the compute requirements but does not change the number of parameters.

\paragraph{Electricity Usage.} A time- and location-agnostic way to quantify training efficiency is the electricity used during DNN training. GPUs and CPUs can report their current power usage, which can be used to estimate the total electricity consumed during training. Unfortunately, electricity usage is dependent on the hardware used for model training, which makes it difficult to perform a fair comparison across methods implemented by different researchers. Moreover, even for fixed hardware, it is possible to trade off power consumption and runtime \citep{you2022zeus}.

\paragraph{Carbon Emission.} The carbon footprint of DNN training is another useful metric. However, accurately measuring carbon emissions can be challenging, requiring inclusion of emissions associated not just with the hardware's use but with its life cycle too \citep{gupta2022chasing,luccioni2022estimating}. Further, running the same training routine twice, consuming the same amount of electricity each time, does not imply the training runs will have the same carbon emissions due to variation in the carbon-intensity of electricity generation. Indeed, this metric depends highly on the local electricity infrastructure and current demand; therefore, one cannot easily compare results when experiments are performed in different locations or even in the same location at different times \citep{schwartz2020green}.

\paragraph{Operand sizes.} The total number of activations in a model's forward pass is a proxy for memory bandwidth consumption and can be a useful proxy for runtime \citep{radosavovic2020designing, dollar2021fast}. This metric can be defined rigorously and independent of hardware for a given compute graph. It also decreases when operators are fused, which may or may not be desirable.

\subsection{Pitfalls of Different Metrics}
\label{sec:pitfalls}

To illustrate how some popular metrics can be misleading, we perform two microbenchmarking experiments. In the first, we profile individual PyTorch \citep{paszke2017automatic} operations on a 40GB A100, including matrix products, factorized matrix products similar to those of \citet{zhang2015accelerating}, and normalization ops. In the second, we profile the time required to perform a forward pass on a batch of 128 $224 \times 224$ images with a ResNet-50. Experimental details and additional experiments can be found in Appendices \ref{app:experiments} and \ref{app:new_experiments}, respectively.

\begin{figure}[!t]
\centering
\includegraphics[width=1\linewidth]{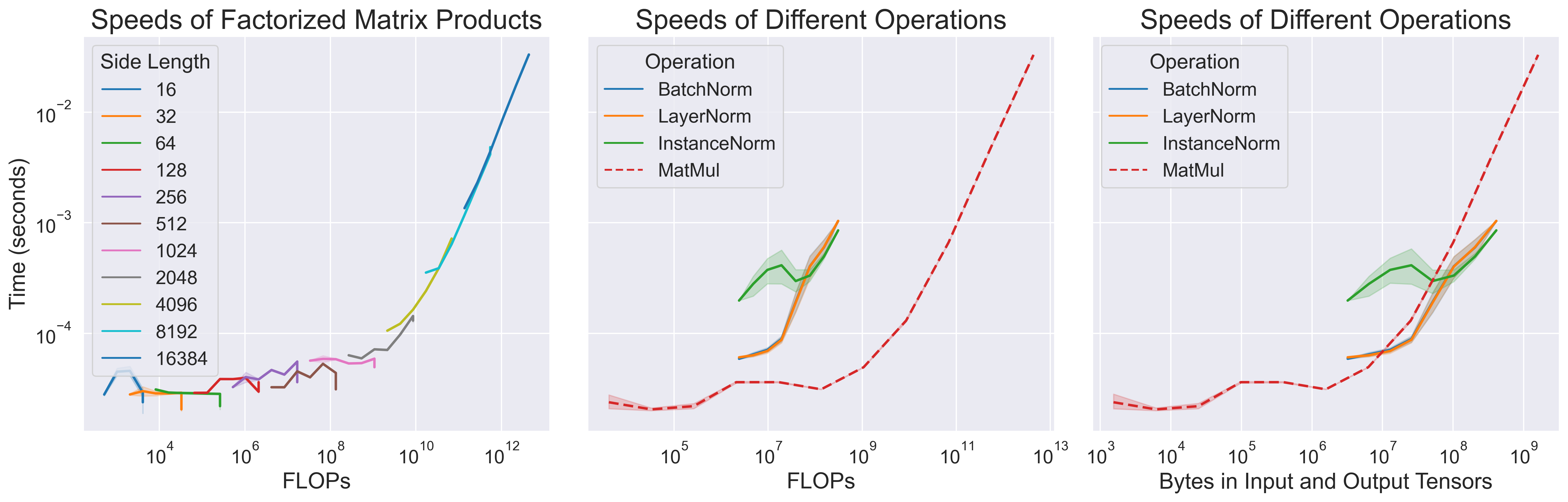}
\caption{Some FLOPs require orders of magnitude more time than others. \textit{(left)} Multiplying matrices of different sizes with different degrees of factorization in the right matrix. Each curve is one size and each point in the curve is one level of rank reduction, with the rightmost point corresponding to a single, full-rank matrix. Small matrix products run in time nearly independent of the FLOP count. Indeed, time and FLOPs are not even monotonically related. \textit{(middle)} For a fixed number of FLOPs, different operations can vary in time by an order of magnitude. Normalization ops, which tax the memory bandwidth far more than the compute units, are much more expensive. \textit{(right)} Total size of input and output operands, a proxy for memory bandwidth consumption, is a more consistent, though far from perfect, predictor of runtime.}
\label{fig:speed_proxies}
\end{figure}

In Figure~\ref{fig:speed_proxies} (left), we see the relationship between FLOP count and time for square matrix multiplies of various sizes. At each size, we also show the results of factorizing the right matrix into two low-rank matrices; this reduces the FLOP count when the rank reduction is greater than two. For matrices of size $\sim4096$ or larger, FLOPs and time are linearly related. But for small matrices, they can be unrelated---the time is more a function of kernel launch overhead, memory bandwidth, and implementation details (the natural explanation for strictly smaller operations taking more time).

In Figure~\ref{fig:speed_proxies} (middle), we similarly see that the relationship between FLOPs and time can vary greatly across operations. Normalization operations, which consume a great deal of memory bandwidth per FLOP, also take much more time per FLOP---as one would expect from an elementary roofline analysis \citep{williams2009roofline}. The operator shapes for the normalization ops are taken from ResNet-50, augmented with channel counts increased and decreased by factors of two for extra coverage. The layer normalization and batch normalization curves coincide because the latter is implemented using the former in PyTorch. Shaded areas indicate standard deviations; these are always present but sometimes tiny due to the log scale.

In Figure~\ref{fig:speed_proxies} (right), we see the same measurements as Figure~\ref{fig:speed_proxies} (middle), but with the total size of all input and output operands used as the x-axis rather than FLOPs. This metric is a more comprehensive alternative to the number of activations that also includes the sizes of parameter tensors. This metric is still far from perfect but eliminates much of the difference across ops.

\begin{figure}[!t]
\centering
\includegraphics[width=1\linewidth]{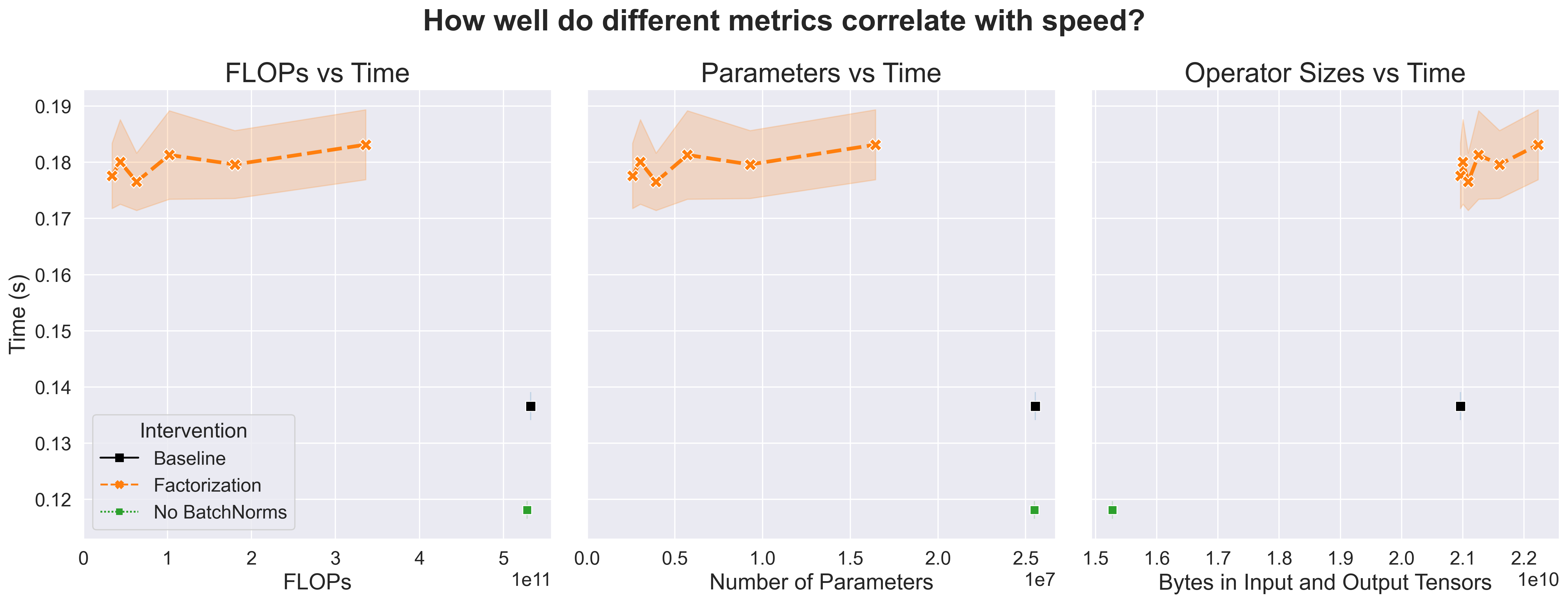}
\caption{FLOP count \textit{(left)} and parameter count \textit{(middle)} are not reliable proxies for wall time when training ResNet-50. They are not even monotonically related to wall time. \textit{(right)} The total size of all operator inputs and outputs is a somewhat better proxy, at least yielding a roughly monotonic relationship. None of these metrics are accurate enough predictors to substitute for measuring time directly.}
\label{fig:rn50_microbench}
\end{figure}

In Figure~\ref{fig:rn50_microbench}, we replicate the results of Figure~\ref{fig:speed_proxies} with an entire ResNet-50, rather than individual operations. Note that accuracy is not held constant in these experiments, as our focus is showing the conflict among efficiency levels suggested by different metrics. As shown in the left subplot, factorizing all of the convolutions and linear layers decreases the FLOP count significantly, but \textit{increases} the runtime thanks to increased memory bandwidth usage and kernel launch overhead. However, removing the batch normalization ops (as is commonly done during inference) reduces time significantly---despite having almost no impact on FLOP count. A similar pattern holds for parameter count in the middle plot.

However, in Figure~\ref{fig:rn50_microbench} (right), we see that measuring the size of input and output operands correctly orders the different ResNet-50 variants---though it is still not a reliable predictor of runtime.
Taken together, these results indicate that the network's operations are largely bottlenecked by memory bandwidth. Removing batch normalization ops reduces data movement, while factorizing linear transforms does not, allowing differences in their speedups.

\begin{figure}[h]
\centering
\includegraphics[width=0.9\linewidth]{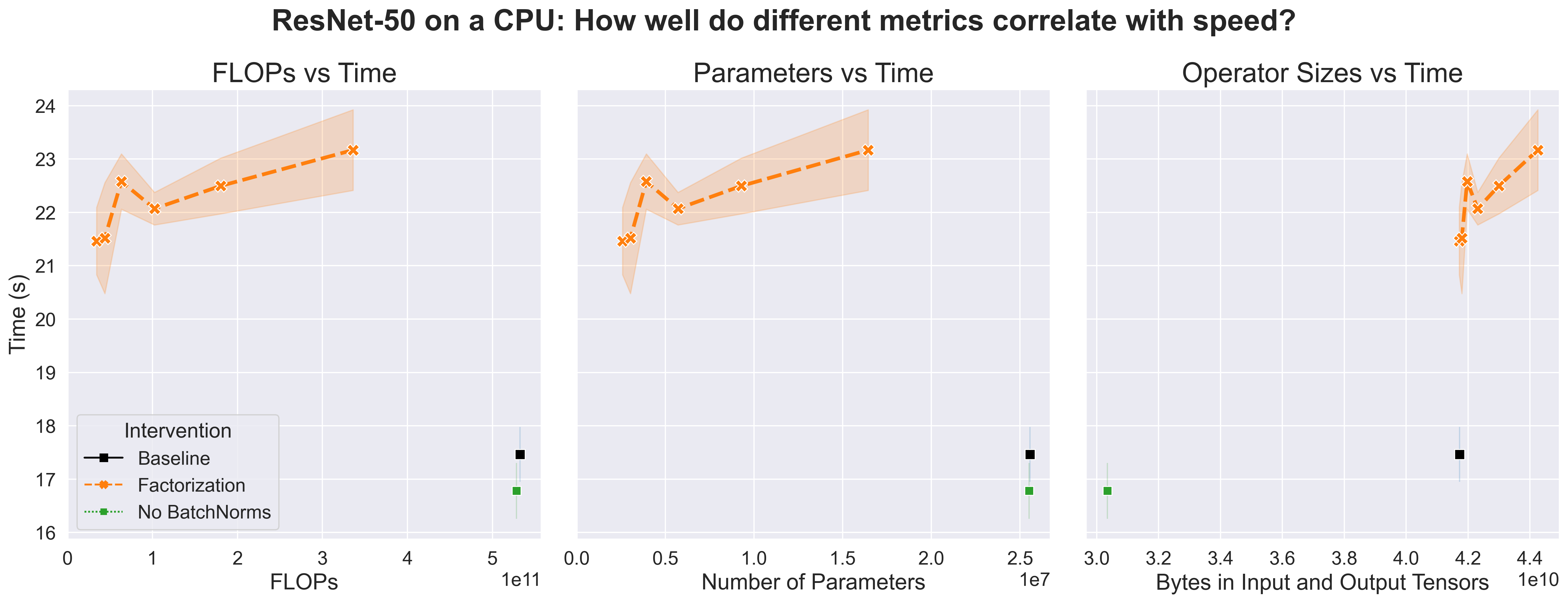}
\caption{Runtime vs various metrics for ResNet-50 on 64 CPU cores.}
\label{fig:resnet50-cpu}
\end{figure}

\begin{figure}[h]
\centering
\includegraphics[width=0.9\linewidth]{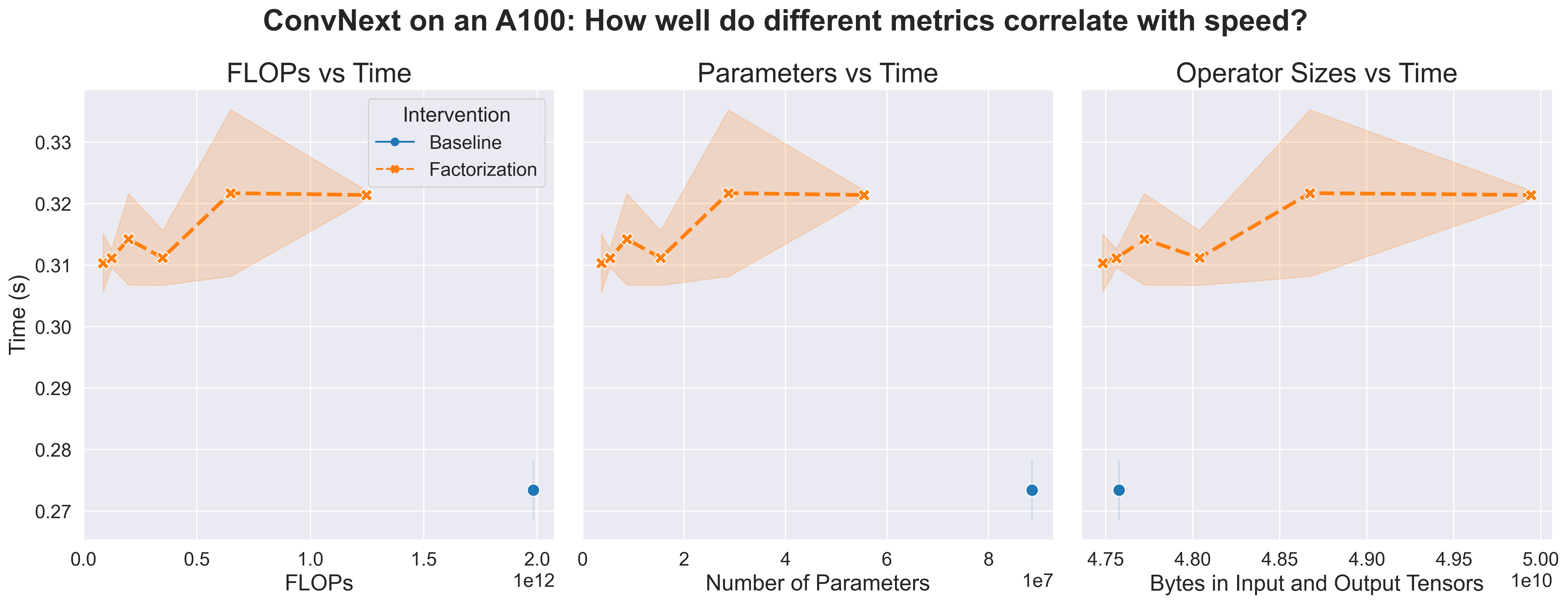}
\caption{Runtime vs various metrics for ConvNeXt-B on an A100 GPU.}
\label{fig:convnext-gpu}
\end{figure}

\begin{figure}[h]
\centering
\includegraphics[width=0.9\linewidth]{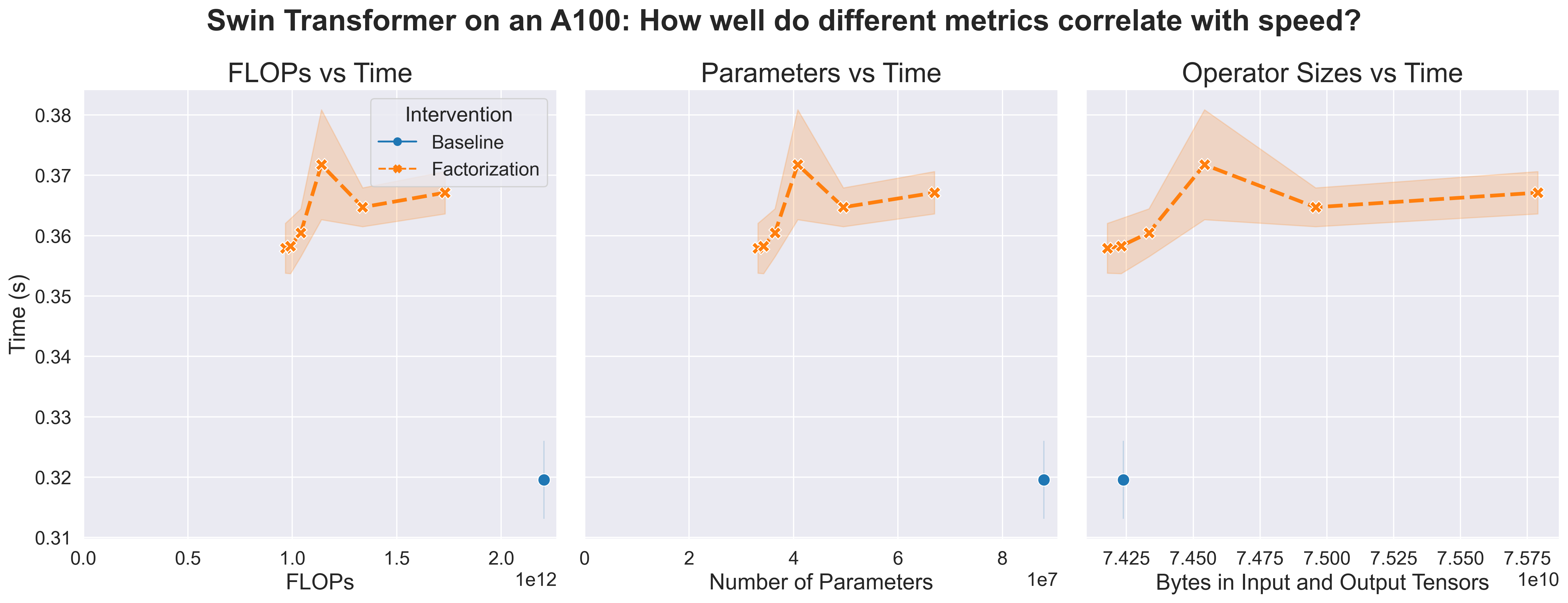}
\caption{Runtime vs various metrics for Swin-B on an A100 GPU.}
\label{fig:swin-gpu}
\end{figure}

This memory bandwidth bottleneck often arises on other hardware and in other networks as well. Illustrating this, we build on Figure~\ref{fig:rn50_microbench} by considering two further variables: hardware and model architecture. Regarding hardware, we measure how various metrics correlate with runtime on a 2-socket server with AMD EPYC 7513 32-core processors to complement our A100 GPU results (see Figure \ref{fig:resnet50-cpu}). Regarding architecture, we collect both CPU and GPU measurements using ConvNeXt-B \citep{convnext} and Swin-B \citep{swin} to complement our ResNet-50 results (CPU results for Swin and ConvNeXt are shown in Appendix \ref{app:new_experiments}). The setup of Figures \ref{fig:convnext-gpu} and \ref{fig:swin-gpu} is identical to that of Figure~\ref{fig:rn50_microbench} with the exception that we omit removal of Batch Normalization as an intervention, since it is not applicable. Just as in Figure~\ref{fig:rn50_microbench}, these results show that a) FLOP count can be a poor proxy for wall time, and b) memory bandwidth consumption can be a better, though still imperfect, proxy.

\ 

\noindent  To summarize, all of the above efficiency metrics have limitations. Either these metrics are a step removed from costs of direct interest or they are dependent on confounding factors such as hardware, location, time, etc. However, by reporting multiple metrics---ideally via a set of accuracy-efficiency curves \citep{portes2022fast}---and holding as many confounding factors as possible constant, one can account for the fact that different metrics may lead to different or even opposite efficiency conclusions and adequately characterize efficiency (as we discuss further in Section~\ref{sec:evaluation}).

{Next, we define algorithmic speedup, adopting training time as our metric of interest. Our definition can be extended to perform training efficiency comparisons in terms of other metrics.}


\subsection{Algorithmic Speedup: Motivation and Definition}

The goal of speedup research is to make training efficient by minimizing the time required to achieve the desired model quality. The most prevalent way to achieve a speedup is to improve software- or hardware-engineering aspects of the training pipeline; e.g., using optimized compute kernels or fusing together adjacent memory-bandwidth-bound operations. However, in this paper, we are interested in a complementary direction that we refer to as \emph{Algorithmic Speedup}---modifying the semantics of the training process to achieve a speedup.
Algorithmic Speedup is formalized as follows.

\begin{definition}[\textbf{($\epsilon$,\;$\delta$)-Speedup}]
    Let $Q$ denote the model quality achieved by the baseline training recipe $\mathcal{R}$ in time $T$. Algorithmic Speedup is an $\epsilon$ reduction in the time required to achieve model quality $Q'$ by algorithmically changing the base recipe to $\mathcal{R}'$, where $\epsilon=\frac{T}{T'}>1$ and $ \delta =\frac{Q'-Q}{Q}$.

\end{definition}

Algorithmic Speedup aims to reduce the resources required to reach a particular model quality via one or both of the following effects: 1) a reduction in the number of iterations required to reach a specific performance level ($n_{\text{iteration}}$); and 2) a reduction in the time per iteration ($T_i$).
If addition of a speedup method to a training recipe improves training time at the same or better model quality, i.e., $(\epsilon>1, \delta\geq 0)$, then the new recipe is strictly better than the baseline. In cases where model quality and training time conflict with each other (i.e., $\epsilon>1, \delta< 0)$, the goal is to find a Pareto optimal solution with a good tradeoff between both objectives. Ranking speedup methods under such scenarios is a challenging task. We discuss this in more detail in Section~\ref{sec:evaluation}.  


\section{A Unifying Perspective for Speedup Methods: Taxonomy}
\label{sec:taxonomy}

\begin{figure}[t]
\centering
\includegraphics[width=.85\linewidth]{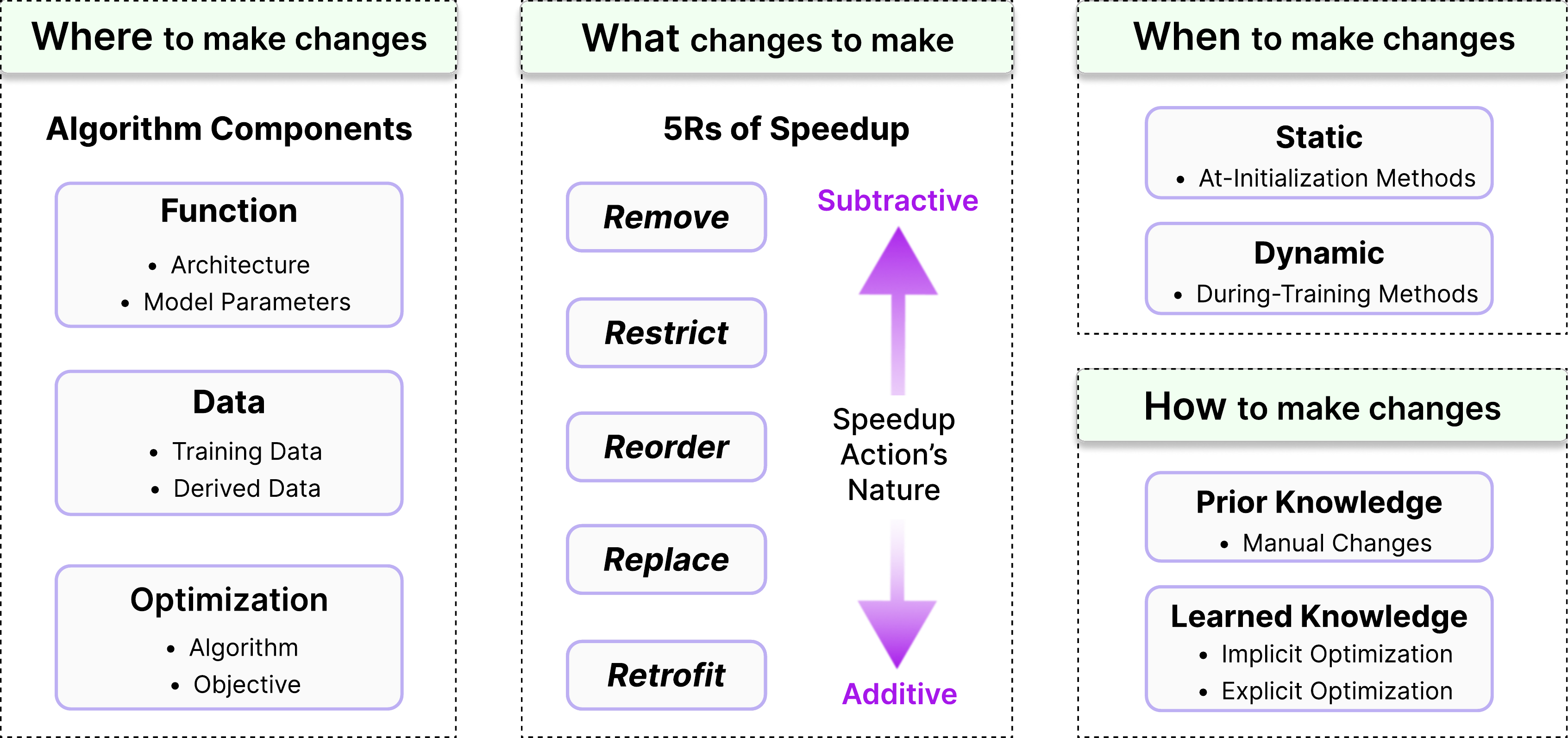}
\caption{\textbf{A Taxonomy of Algorithmic Speedup Techniques.} Our taxonomy classifies algorithmic speedup approaches according to the four analysis questions above. We identify three fundamental building blocks (i.e., component, action, and mechanism) and present the subcomponents that the components comprise. Any speedup technique can be viewed as a path connecting different elements across these blocks. Note that multiple elements can be chosen from component, action, and mechanism blocks. New approaches can be derived by changing the elements and/or connecting a different path.}
\label{fig:taxonomy}
\end{figure}

In this section, we introduce a taxonomy that categorizes algorithmic speedup approaches based on three fundamental building blocks of speedup methods (Figure~\ref{fig:taxonomy}):
\begin{enumerate}
    \item \textbf{Components:} Where to make changes?
    \item \textbf{Actions:} What changes to make?
    \item \textbf{Mechanisms:} When and how to make changes?
\end{enumerate}

A detailed categorization of the literature according to this taxonomy with concrete examples will be presented in the next section (Section~\ref{sec:survey}).
Here we describe the taxonomy on a more conceptual level.

\subsection{Component Categorization}
The \emph{component} building block indicates the aspect of the training pipeline to be modified (see Figure~\ref{fig:pipeline}). Namely, the function, the data, or the optimization.
Each component includes subcomponents, such as architecture and model parameters, which we list below.

\subsubsection{Function Component}
\vspace{0.1in}

\begin{itemize}
    \item \textbf{Architecture:} A neural network architecture defines the structure of the model and encodes inductive biases about the problem at hand. Two examples of neural network architectures are ResNet-50 \citep{he2016deep} and DenseNet \citep{huang2017densely}. An architecture is composed of many layers, which we regard as elements of the architecture subcomponent. 
    \item \textbf{Model Parameters:} Model parameters are the free variables of a given architecture, which are trained to maximize its predictive performance.
\end{itemize}

\subsubsection{Data Component}
\vspace{0.1in}

\begin{itemize}
    \item \textbf{Training Data:} A training dataset is a collection of data samples used during the learning process to fit the model parameters. A data sample is usually a pairing of a raw data instance (e.g. image, text, or audio) and a corresponding annotation (e.g. a label or bounding boxes).  Factors such as the number of samples and the size of each sample (e.g. resolution or sequence length) can significantly impact the training time.
    \item \textbf{Derived Data:} Derived data is data that is computed from training data during the learning process. Derived data appears during the forward and backward passes of the DNN. Some examples are activation values in the forward pass and gradients in the backward pass.
\end{itemize}

\subsubsection{Optimization Component}

\begin{itemize}
    \item \textbf{Training Objective:} Neural networks are trained by optimizing an objective function computed using training data. An objective function is often divided into a loss function (e.g., cross-entropy or hinge loss) and a regularization penalty (e.g., weight norm or sharpness). 
    It is usually desirable for objective functions to be easy to optimize, meaning that they are ideally smooth, as convex as possible, and well-conditioned.
    \item \textbf{Training Algorithm:} Training algorithms modify the model parameters to optimize the training objective. The training algorithm is composed of the initializer (e.g., Xavier or  Kaiming initialization) and the optimizer (e.g., SGD or Adam). Initialization sets the model parameters to certain initial values that define the starting point for the optimizer. Optimizers are methods of changing the parameters to optimize the objective function, usually based on the gradient of the objective with respect to the parameters. Arguably more sensitive than other subcomponents to hyperparameter settings, the training algorithm can cause training to fail entirely if inappropriately configured. 
\end{itemize}

\subsection{Action Categorization}
The \emph{action} building block indicates the type of operation applied to a given component to achieve a speedup. 
Speedup methods tend to use one of the following five actions, which we call the \textbf{\emph{5Rs of algorithmic speedup}}. 

\subsubsection{Remove}
\begin{figure}[t]
\centering
\includegraphics[width=.9\linewidth]{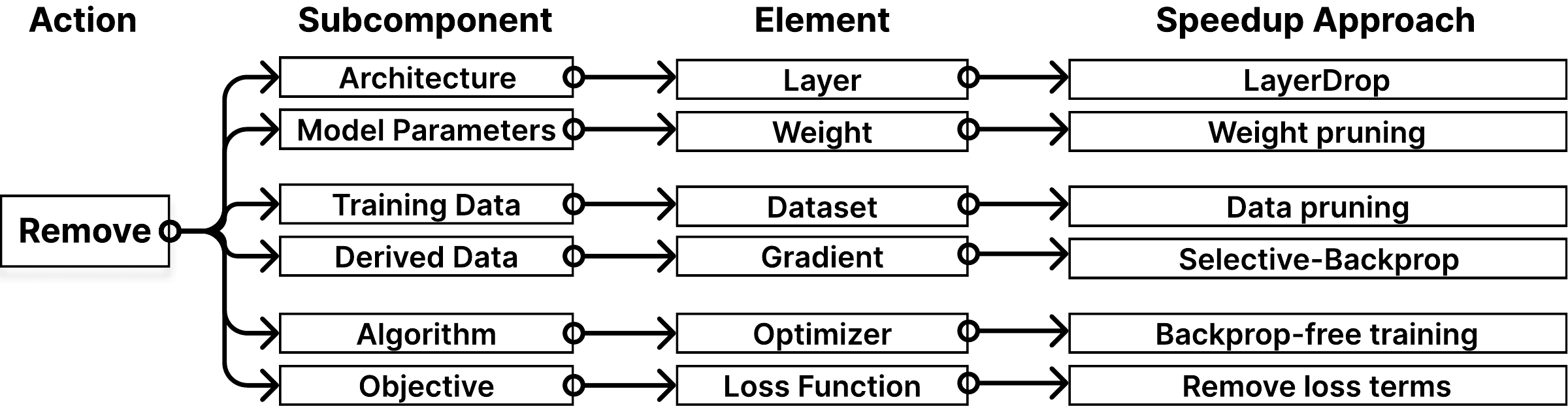}
\caption{Illustration of example ``remove'' speedup techniques inside our taxonomy.}
\label{fig:remove}
\end{figure}

This category of actions is subtractive, removing elements of the components they target. The remove operation typically aims to reduce the $T_{\text{total}}$ needed to attain the desired accuracy by reducing $T_i$.
Representative ``remove'' techniques for each subcomponent are illustrated in Figure \ref{fig:remove}.

\subsubsection{Restrict}
\begin{figure}[t]
\centering
\includegraphics[width=.9\linewidth]{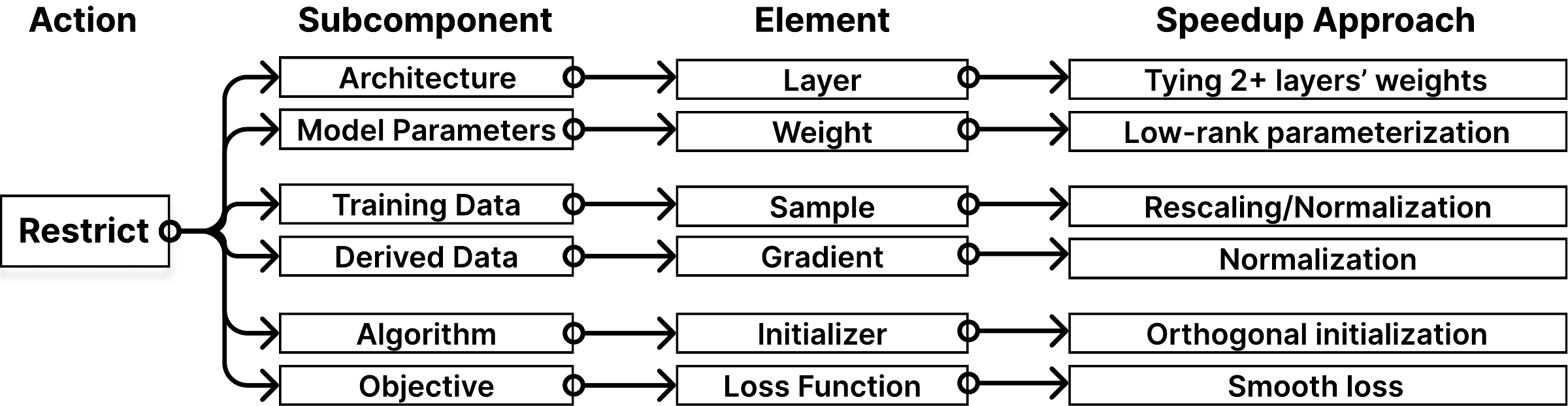}
\caption{Illustration of example ``restrict'' speedup techniques inside our taxonomy.}
\label{fig:restrict}
\end{figure}

Components can often take on a range of possible values---e.g., all of $\mathcal{R}^D$ for a $D$-element parameter tensor. The restrict action shrinks the space of possible values in some way. Actions in this category typically aim to reduce the $T_{\text{total}}$ needed to attain the desired accuracy by reducing $T_i$ but may also reduce $n_{\text{iteration}}$. 
Representative ``restrict'' techniques for each subcomponent are illustrated in Figure \ref{fig:restrict}.

\subsubsection{Reorder}
\begin{figure}[t]
\centering
\includegraphics[width=.9\linewidth]{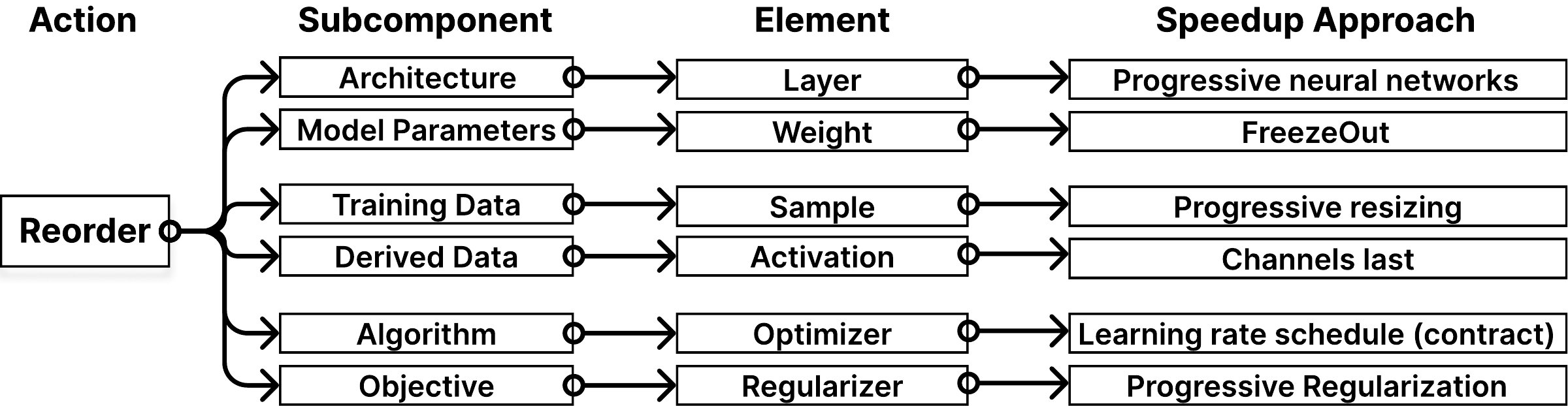}
\caption{Illustration of example ``reorder'' speedup techniques inside our taxonomy.}
\label{fig:reorder}
\end{figure}

Without adding to or subtracting from its elements, training can be sped up by altering the places and phases in which its elements are introduced and used.
Reorder techniques can reduce $T_i$ through progressively increasing problem complexity. They can also reduce $n_{\text{iteration}}$ by, e.g., reordering samples to improve optimization.
Representative ``reorder'' techniques for each subcomponent are illustrated in Figure \ref{fig:reorder}.

\subsubsection{Replace}
\begin{figure}[t]
\centering
\includegraphics[width=.9\linewidth]{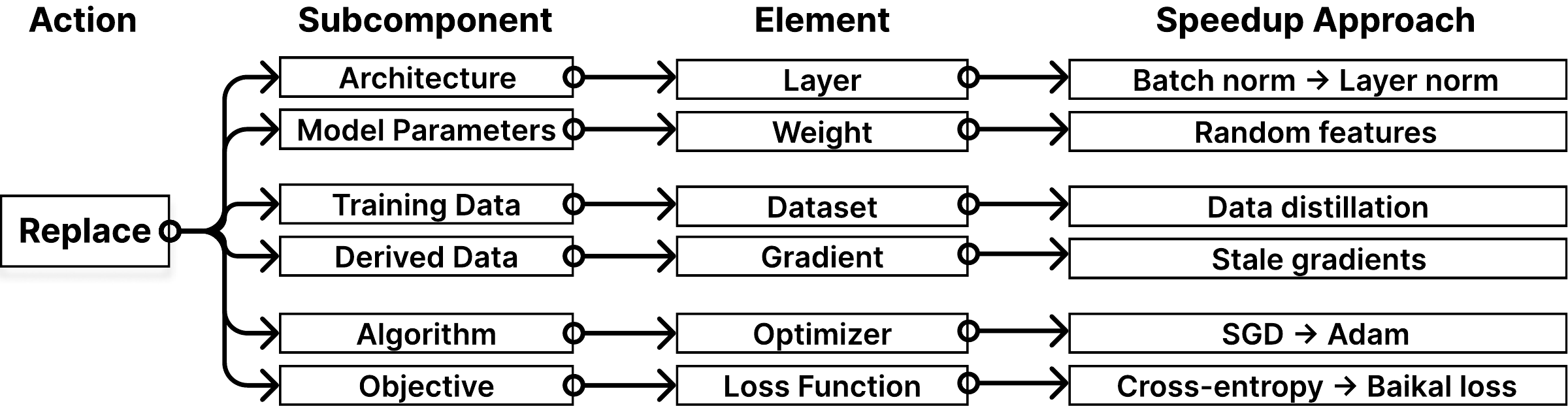}
\caption{Illustration of example ``replace'' speedup techniques inside our taxonomy.}
\label{fig:replace}
\end{figure}

Sometimes one can obtain a speedup by completely replacing one subcomponent/element with another.
This can reduce both $T_i$ and $n_{\text{iteration}}$.
Representative ``replace'' techniques for each subcomponent are illustrated in Figure \ref{fig:replace}.

\subsubsection{Retrofit}
\begin{figure}
\centering
\includegraphics[width=.9\linewidth]{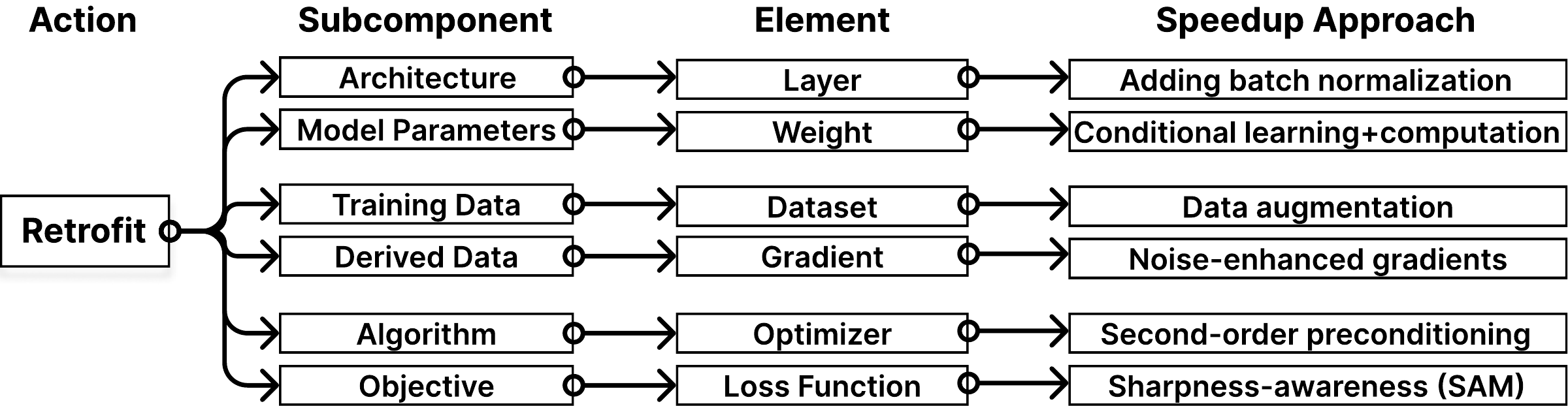}
\caption{Illustration of example ``retrofit'' speedup techniques inside our taxonomy.}
\label{fig:retrofit}
\end{figure}

Retrofitting adds to components and is the opposite of the remove action.
Because it typically does not reduce work done per iteration, retrofitting aims to reduce $n_{\text{iteration}}$ rather than $T_i$.
Representative ``retrofit'' techniques for each subcomponent are illustrated in Figure \ref{fig:retrofit}.

\subsection{Mechanism Categorization}
The \emph{mechanism} building block describes when and how to perform an action on a component. For ``when,'' we consider time in relation to the training process. For ``how,'' we focus on what information determines the details of the change.

\subsubsection{When to Make the Change?}
\begin{itemize}
    \item \textbf{Static Methods:} These methods perform component-action changes only once at initialization, i.e., before the training starts. 
    \item \textbf{Dynamic Methods:} These methods perform component-action changes during training, e.g., at various training iterations. 
\end{itemize}
 
\subsubsection{Change Based on What Information?}
Component-action changes can be made based on various sources of knowledge. We categorize speedup methods based on: 
\begin{itemize}
    \item \textbf{Domain-Knowledge-Based Methods:} These methods perform component-action changes using domain knowledge, i.e., prior experience or expertise related to the training process. 
    \item \textbf{Learning-Based Methods:} These methods perform component-action changes by either explicitly or implicitly optimizing for faster training time via leveraging the data corresponding to current (or past) training runs.
\end{itemize}

\begin{table}
\centering
{\setstretch{0.75} \fontsize{7pt}{10.25pt}\selectfont{
\begin{tabular}{p{.13\textwidth}p{.08\textwidth}p{.09\textwidth}p{.3\textwidth}p{.3\textwidth}}
    \toprule
    {\scriptsize Subcomponent} &
    {\scriptsize Action} &
    \scriptsize{Timing of Action} &
    \scriptsize{Domain-Knowledge Based} &
    \scriptsize{Learning Based }
    \\
    \midrule
    \multirow{4.5}{*}{\hfil \shortstack[l]{\scriptsize{Model} \\ \scriptsize{parameters}  }}
    &
    \multirow{4.5}{*}{\hfil Remove }
    &
    \multirow{2}{*}{\hfil Static }
    &
    Sparsification at initialization using magnitude pruning
    &
    Sparsification at initialization using meta-learning
    \\
    \cmidrule{3-5}
    
    &
    
    &
    \multirow{2}{*}{\hfil Dynamic }
    &
    Dynamic sparsification using gradual magnitude pruning
    &
    Dynamic sparsification using score-based optimization
    \\
    \midrule
    \multirow{4.5}{*}{\hfil \shortstack[l]{\scriptsize{Training} \\ \scriptsize{data}  }}
    &
    \multirow{4.5}{*}{\hfil Remove }
    &
    \multirow{2}{*}{\hfil Static }
    &
    Coreset selection at initialization using expert knowledge
    &
    Coreset selection at initialization using meta-learning
    \\
    \cmidrule{3-5}
    
    &
    
    &
    \multirow{2}{*}{\hfil Dynamic }
    &
    Minibatch selection using expert-guided sampling
    &
    Adaptive subset selection using bilevel optimization
    \\
    \midrule
    \multirow{4.5}{*}{\hfil \shortstack[l]{\scriptsize{Training} \\ \scriptsize{algorithm}  }}
    &
    \multirow{4.5}{*}{\hfil Replace }
    &
    \multirow{2}{*}{\hfil Static }
    &
    Fixed learning rate found by experts (or theory)
    &
    Fixed learning rate predicted using regression or range test 
    \\
    \cmidrule{3-5}
    
    &
    
    &
    \multirow{2}{*}{\hfil Dynamic }
    &
    Learning rate schedule found by experts (or theory)
    &
    Adaptive learning rates predicted using regression 
    \\
    \bottomrule
\end{tabular}
}}
\caption{
    \textbf{Illustrative Examples of Speedup Mechanisms.} These examples show how and when speedup actions can be applied.
}
\label{tab:how_table}
\end{table}

Some examples of mechanisms for popular subcomponent-action combinations are provided in Table~\ref{tab:how_table}.
Finally, we note that some speedup methods can be associated with more than one of the paths through our taxonomy (e.g., a method might be reasonably classified as both a \textbf{replace} $\rightarrow$ \textbf{derived data} and a \textbf{replace} $\rightarrow$ \textbf{training algorithm} method). This stems from the fact that our taxonomy was not designed to be a mutually exclusive and collectively exhaustive categorization of the literature. Instead, our taxonomy aims to group speedup methods to aid reasoning about their potential effectiveness given a particular bottleneck, as well as reveal themes and opportunities among the broad set of approaches we review.

\section{Categorization of Existing Work: Survey}
\label{sec:survey}

Here, we discuss the representative literature on algorithmic speedup methods. 

\subsection{Function Speedup Strategies} \label{sec:funcSpeedups}

Function speedup strategies apply the 5Rs to the model parameters and architecture.
Function speedup strategies that reduce function latency like \textbf{remove} $\rightarrow$ \textbf{model parameters} (e.g., pruning) can reduce $T_i$ when the function is a bottleneck. Alternatively, more additive speedup strategies like \textbf{retrofit} $\rightarrow$ \textbf{architecture} (e.g., widening layers) tend to help reduce $n_{\text{iteration}}$.
We place various function speedup options into our taxonomy and review them below.

\subsubsection{Model Parameters}
\label{sec:model_param}


\paragraph{Remove} \textbf{Remove} $\rightarrow$ \textbf{model parameters} is also known as model pruning \citep{hoefler2021sparsity} and has been approached in many ways. We focus our discussion on several examples that explicitly demonstrate training speedups despite the fact that sparse matrix computations do not offer significant benefits on most AI accelerators.\footnote{Emerging hardware like \citet{cerebras2021stream} may accelerate unstructured sparsity in the future.} In particular, we discuss approaches that attain speedups by pruning structures larger than individual weights \citep{lym2019prunetrain, chen2020earlybert, yuan2021mest}.
Structured pruning entails removing a block of parameters, rather than individual parameters. E.g., one might remove an entire filter from a convolutional layer. For some block structures, this can produce speedups without specialized hardware \citep{yuan2021mest}.

Pruning-based acceleration should also facilitate accuracy similar to the unpruned baseline model's, which is difficult to reach when training from scratch with a fixed and naively-applied sparsity pattern \citep{liu2018rethinking,frankle2020pruning}. \citet{lym2019prunetrain} address this with their method PruneTrain, which applies group lasso regularization to filters/channels in convolutional networks to encourage the emergence of a sparsity pattern during training. PruneTrain periodically computes the maximum absolute value of the parameters in each filter, then removes the filters with the smallest values to reduce $T_i$. \citet{lym2019prunetrain} shows this approach produces speedups without a significant loss of accuracy.
The intuition behind the success of this approach is that once weights are sparsified by group lasso, they rarely grow above the threshold later in training. Thus, these weights can be pruned without degrading accuracy. 

A similar regularization-based approach is taken by EarlyBERT \citep{chen2020earlybert} to accelerate pretraining and fine-tuning of large language models. Specifically, \citet{chen2020earlybert} encourage sparsity by applying $\ell_1$ regularization to coefficients on outputs associated with groups of parameters in a manner inspired by \citet{liu2017learning}. A key difference is that the parameter groups in language models are neurons in fully connected layers and attention heads rather than convolutional filters. \citet{chen2020earlybert} also take inspiration from the Early-Bird ticket algorithm \citep{you2019drawing}, which was shown to reduce the energy cost of training by pruning early in the training process without sacrificing accuracy. \citet{you2019drawing} use the scaling factors in batch normalization layers as indicators of the corresponding channels' significances.
Early-Bird tickets are identified ``early'' when the Hamming distance between the pruning masks on successive iterations becomes sufficiently small \citep{you2019drawing}. EarlyBERT \citep{chen2020earlybert} uses Hamming distance during training to justify its early selection of a sparsity pattern as well, 
and it enforces sparsity before 10\% of the total training iterations are reached.
They find that this not only reduces $T_i$ but also $n_{\text{iteration}}$, thanks to an increased accuracy early in training.

\citet{yuan2021mest} address the issue of picking a suboptimal initial sparsity mask by using a dynamic sparsity mask throughout training in their MEST pruning framework. 
During training, MEST periodically runs a prune-and-regrow iteration that removes parameters with both small weight and gradient magnitudes, then reactivates previously removed parameters such that the model is always trained with a constant sparsity level. MEST's dynamic approach facilitates a low memory footprint via sparse gradient and parameter vectors, while not requiring the permanent removal of potentially important connections. Notably, since pruning larger structures is associated with worse accuracy, \citet{yuan2021mest} also explore pruning structures of various sizes to find a good balance between accuracy and speedup.

\citet{adelman2021faster} use sample-based approximations to tensor operations---i.e. matrix multiplications and convolutions. This involves dynamic row/column pruning of the tensors in these operations. On ImageNet, models trained with this sampling can train $1.37\times$ faster with little impact on the final test accuracy.

Learning-based approaches dominate \textbf{remove} $\rightarrow$ \textbf{model parameters} due to the difficulty of reaching good accuracy with a predefined or random sparsity pattern. The development of domain-knowledge approaches could remove the computational overhead of discovering a good sparsity pattern through experimentation. This may involve deepening our understanding of initializing, regularizing, and/or selecting the connectivity of sparse networks. 


\paragraph{Restrict} There are multiple ways to \textbf{restrict} $\rightarrow$ \textbf{model parameters} for speedup, including both weight quantization and low-rank factorization methods. Similar to parameter removal, parameter restriction is a straightforward way to reduce theoretical FLOPs. However, realizing these gains while maintaining accuracy is challenging.

Restricting a weight matrix to be less than full rank by factorizing it into two smaller matrices can reduce $T_i$ when computations associated with the weight matrix can be performed faster with its two small factors. This often requires the weight matrix to be relatively large. \citet{sainath2013low} show that applying this low-rank factorization to the large, densely-connected output layer in acoustic and language models can produce speedups without significantly harming accuracy.  While low-rank approaches often focus on such fully connected layers, \citet{ioannou2015training} show that factorizing convolutional filters in VGG-11 and similar networks can also produce speedups without significantly harming accuracy. \citet{ioannou2015training} develop a variety of factorization schemes, but a core idea is to approximate $w \times h$ filters as a layer of $w \times 1$ and $h \times 1$ basis filters followed by a layer of $1 \times 1$ filters that linearly combines the outputs of the basis-filter layer.

The Pixelated Butterfly \citep{chen2021pixelated} (or Pixelfly) approach can translate FLOPs reductions from restrictions into speedups when models are dominated by sufficiently large General Matrix Multiply (GEMM) operations. Pixelated Butterfly applies the following reparameterization of the weight matrix $W$ before training begins: $W = \gamma B + (1-\gamma) UV^T$, where B is a block sparse variant of a butterfly matrix, $UV^T$ is a low-rank component, and $\gamma$ is learnable. \citet{chen2021pixelated} prove that this combination of sparse and low-rank matrices is more expressive than using just a sparse or just a low-rank matrix. Moreover, \citet{chen2021pixelated} find that applying Pixelated Butterfly to various Transformer and MLP-mixer models leads to high accuracies along with speedups. Similarly, \citet{dao2022monarch} propose Monarch matrices that are hardware-efficient (i.e., parameterized as products of two block-diagonal matrices for better hardware utilization) and expressive (i.e., can represent many commonly used transforms).

Indeed, the sizes of large transformer models have motivated a variety of training efficiency approaches. \citet{hu2021lora} provide another example of restricting transformer parameters via their LoRA algorithm, which facilitates fast and high-quality fine-tuning of pretrained language models by learning low-rank update matrices that are added to the pretrained weight matrices (avoiding fine-tuning of the full weight matrices). Note that we discuss even more efficient transformer approaches, including low-rank-like approaches related to efficient transformer \textit{attention}, in Section \ref{sec:derived}.

Quantization is another way to restrict parameters and is commonly used to reduce inference cost---see \citep{gholami2021survey} for a survey of such techniques. However, \citet{micikevicius2017mixed} show that using a mix of half and single-precision weights and/or activations at training time (``mixed precision training'') can produce large speedups, particularly on newer AI accelerators that specifically support fast half-precision computation. Notably, \citet{micikevicius2017mixed} avoid accuracy reductions associated with training with lower precision weights in part by avoiding zeroing of parameter gradient information, which is accomplished via 1) scaling up the loss so that small-magnitude gradients are pushed into a range that is captured by half-precision weights (``loss scaling''); and 2) representing the half-precision gradient in single precision before multiplying it by the step size. Many modern deep learning frameworks provide automatic mixed precision (AMP) training functionality that addresses these subtleties for users. 

In terms of mechanisms, \textbf{restrict} $\rightarrow$ \textbf{model parameters} approaches for fast training are mostly static---e.g., the rank is chosen prior to training. Interestingly, there exist more dynamic approaches that can find an appropriate rank for each layer \citep{idelbayev2020low}, but our review did not find an example of such a method that also provides a speedup (with \citeauthor{liebenwein2021compressing}, \citeyear{liebenwein2021compressing}, being a potential exception for pipelines involving retraining). This is due to the added cost of dynamically finding the appropriate ranks of the parameter matrices. Reducing this cost could lead to low-rank methods that adapt to training dynamics, which might improve their efficacy.

\paragraph{Reorder}
Speedups that \textbf{reorder} $\rightarrow$ \textbf{model parameters} mostly exploit the observation that parameters in some layers train faster than those in other layers \citep{raghu2017svcca}. This creates an opportunity to gradually freeze parameters, rather than train all parameters all the time. When a parameter is ``frozen,'' it no longer requires gradient or parameter update computations.

\citet{brock2017freezeout} propose FreezeOut to reduce the training time by only training each layer for a set portion of the training schedule, progressively ``freezing out'' layers and excluding them from the backward pass. Specifically, cosine annealing without restarts is used in a layer-wise manner such that the set of trained model parameters shrinks each time a layer's learning rate is annealed to zero.
Similarly, \citet{raghu2017svcca}
introduce ``Freeze Training'' to successively freeze lower layers during training.

In terms of mechanisms, using progressive training to \textbf{reorder} $\rightarrow$ \textbf{model parameters} is dynamic by definition. Moreover, most of these approaches use knowledge-based mechanisms. An exception is PipeTransformer, proposed by \citet{he2021pipetransformer}, which uses a learning-based mechanism. PipeTransformers 
use a gradient-norm-based objective to identify and freeze layers gradually during training. The resulting scheme is lightweight and achieves up to a 2.83x speedup without losing accuracy. We accordingly suspect that there is great potential for other learning-based approaches in this area.


\paragraph{Replace} \textbf{Replace} $\rightarrow$ \textbf{model parameters} is made difficult by the need to match the replacement parameters to the learning task at hand. This matching process is illustrated by \citet{sarwar2017gabor}, who note that Gabor filters are a powerful processing tool in computer vision and replace a fraction of convolutional weight kernels with fixed Gabor filters. These fixed filters do not need to be trained, which facilitates improved training speed. High accuracy is reached due to the filters' suitability for computer vision applications \citep{sarwar2017gabor}. 

Similarly, \citet{shen2020reservoir} build on the observation that transformers can perform well with random attention by introducing ``Reservoir Transformers”. Specifically, \citet{shen2020reservoir} replace a subset of a transformer's layers with random non-linear transformations.
They show that replacing trainable layers with these random reservoir layers improves both speed and generalization.

From a mechanism perspective, it's clear that the approaches we reviewed are more static in that \textbf{replace} $\rightarrow$ \textbf{model parameters} happens only once, prior to training's start. Additionally, the reviewed approaches rely on domain knowledge regarding the suitability of the replacement for the new task. A potential opportunity is applying replacement operations more dynamically based on learning. For example, a technique might find speedups by choosing the most appropriate replacement for a task according to the performance of a variety of candidates on a subset of that task.

\paragraph{Retrofit} Adding model parameters to lower $n_{\text{iteration}}$ via increased model capacity is typically achieved via architecture changes, but there are creative ways to \textbf{retrofit} $\rightarrow$ \textbf{model parameters} without requiring significant architectural changes. For example, \citet{fedus2021switch} retrofit a transformer's baseline parameter set with additional parameters to create Switch Transformers, which have up to a trillion parameters but the ability to train seven times faster than a T5 transformer. By using subsets of its larger parameter set conditionally, Switch Transformers are able to have the same basic architecture and computations as a baseline transformer model despite having many more parameters. Notably, Switch Transformers employ a switching/routing mechanism to select the specific subset of parameters to use based on the input data \citep{fedus2021switch}. The parameter-enhanced model is more sample efficient, which means it has lower training times due to the lower $n_{\text{iteration}}$ required to reach a given model quality. 

Alternatively, \citet{izmailov2018averaging} add a new set of parameters that are the average of multiple points along the SGD-trained-parameter trajectory in their Stochastic Weight Averaging (SWA) method. They show SWA leads to better model quality per parameter update and has little overhead, making speedups possible through reduced $n_{\text{iteration}}$. \citet{yang2019swalp} extend SWA to low-precision training. Relatedly, \citet{kaddour2022stop} propose recording the state of the model parameters at the end of each training epoch in their LAtest Weight Averaging (LAWA) algorithm. At the start of each epoch, the model parameters are defined as the average of the last $k$ prior parameter states, and \citet{kaddour2022stop} show this leads to significant speedups. For parallel training, \citet{li2022branch} introduce the Branch-Train-Merge (BTM) approach, which averages copies of a large language model (each trained on its own intelligently-selected data subset), facilitating faster distributed training and better accuracy.

Using a load-balancing loss term to ensure the conditionally used parameters are well utilized and trained, \citet{fedus2021switch} illustrate the importance of learning-based mechanisms to \textbf{retrofit} $\rightarrow$ \textbf{model parameters}. \citet{fedus2021switch} also provides (for the first time, to the best of our knowledge) an empirical study of parameter count scaling that doesn't affect FLOPs. This emerging domain knowledge and knowledge regarding benefits of weight averaging \citep{kaddour2022questions} may each be useful for developing new \textbf{retrofit} $\rightarrow$ \textbf{model parameters} approaches.

\subsubsection{Architecture} \label{sec:architecture}


\paragraph{Remove} \textbf{Remove} $\rightarrow$ \textbf{architecture} often immediately reduces $T_i$ through the removal of entire layers. However, work in this area must take steps to maintain performance and/or adapt other aspects of the network to account for the removal. For instance, batch normalization layers have a large computational cost but are important to model performance, providing both regularizing and stabilizing effects. Accordingly, 
to remove the batch normalization layers, \citet{brock2021high}
take several steps to regularize and stabilize the training in other ways. They regularize by adding Dropout \citep{srivastava2014dropout}, Stochastic Depth \citep{huang2016deep}, and a variety of data augmentations. To improve stability, they scale down residual branch outputs, standardize the weights, and (critically for the large batch size and learning rate regime) adaptively clip the gradients based on the scale of the parameters \citep{brock2021high}.

\citet{huang2016deep,fan2019reducing} show that networks can be resilient to the removal of entire blocks of computation (sets of layers) when the removal is applied stochastically throughout training. Specifically, these approaches effectively apply Dropout \citep{srivastava2014dropout}, but the structures that are dropped are ResNet blocks \citep{huang2016deep} or transformer layers \citep{fan2019reducing} rather than neurons. \citet{huang2016deep} show that good performance is attained by their Stochastic Depth method when the drop probability for earlier ResNet blocks is lower than that of later layers, based on the intuition that the extraction of low-level features should reliably be performed on each forward pass. Alternatively, \citet{fan2019reducing} show good performance of their LayerDrop method when using a constant drop probability for all transformer layers. Each approach reduces $T_i$ and can actually reduce the $n_{\text{iteration}}$ required to attain a desired performance level through the regularization effect that random layer dropping provides \citep{huang2016deep, fan2019reducing}.

Our review of popular \textbf{remove} $\rightarrow$ \textbf{architecture} methods suggests that such approaches have a strong dependence on domain knowledge, requiring an understanding of how to replace the effects of removed layers (for example), and tend to be dynamic. Interestingly, while the removal of batch normalization is a static action that NFNets use to attain speedups, \citet{brock2021high} compensated for the loss of batch normalization by dynamically altering gradients/weights and adding regularization throughout training.

\paragraph{Restrict} 
Our review found \textbf{restrict} $\rightarrow$ \textbf{architecture} for speedups approached in only one way, which may reflect the difficulty of developing ways to restrict entire layers of architectures.

\citet{press2016using} do this by
tying the input embedding and output embedding matrices of language models.
They show that restricting embedding layers in this way results in better perplexity per training iteration, which means this technique lowers $n_{\text{iteration}}$. Weight tying is also used in modern Transformer-based language models, where it can save even more embedding parameters thanks to the presence of both encoder and decoder embedding matrices \citep{vaswani2017attention}.


\paragraph{Reorder} While it may seem challenging to find inefficiencies in the ordering of popularly used architectures, our survey revealed multiple ways to achieve speedups with \textbf{reorder} $\rightarrow$ \textbf{architecture}. For example, \citet{saharia2022photorealistic} design a faster U-Net (Efficient U-Net) for the Imagen text-to-image model by applying U-Net's downsampling (upsampling) \textit{before} (\textit{after}) the convolutional layers in the downsampling (upsampling) blocks. This approach causes the convolutional layers to operate on smaller feature maps than they would if the original order were used, which improves training speed via reduced $T_i$ without quality degradation.

Separately, many progressive learning approaches reorder the layers of smaller, quickly-trained (or pretrained) networks to initialize a large network that can be trained faster than its randomly initialized counterpart. For example, \citet{chen2015net2net} proposed Net2Net, which accelerates the training of a large network by building it from the pretrained layers of a network with less depth/width. The strategy for growing the network's depth/width ensures that the grown network's output matches the smaller network's output when training starts. This enables a reduced $n_{\text{iteration}}$ relative to using a randomly-initialized large model.
Similarly, BERT pretraining can be accelerated via training progressively deeper BERT models created by the stacking of shallower BERT models' layers \citep{gong2019efficient} or growing in multiple network dimensions \citep{gu2020transformer}.

Notably, these methods build architectures using regular reorderings of fundamental components, such as by duplicating each layer to increase the depth by a factor of two. This regularity might be suboptimal on a broader set of learning tasks, which motivates the automated progressive learning scheme AutoProg \citep{li2022automated}.
During training, AutoProg repeats the following three steps to gradually build a ViT with no accuracy drop relative to the large ViT baseline: 1) train the best-performing ViT subnetwork found so far; 2) scale the trained subnetwork into an Elastic Supernet, and 3) train a variety of subnetworks sampled from this Elastic Supernet. \citet{li2022automated} connect this approach to their ``Growing Ticket Hypothesis of ViTs'', which states that the performance of a large ViT model can be reached in the same number of training iterations by instead progressively growing a ViT from one subnetwork. Due to the lower $T_i$ associated with training subnetworks instead of the large ViT and the constant $n_{\text{iteration}}$, this approach can speed up ViT ImageNet training by 40-85\%.

A related but slightly different direction is training each layer or block of layers individually rather than at the same time, which frees up GPU memory. \citet{huang2018learning} provide some evidence that this can improve training time not only through a reduced number of required steps to reach a given performance level but also through faster step time. However, the authors also show this approach can sometimes lead to worse model quality relative to end-to-end optimization. 

Mechanistically, the \textbf{reorder} $\rightarrow$ \textbf{architecture} approaches we found are mostly domain knowledge, dynamic approaches.
However, there is some evidence \citep{li2022automated} that using learning rather than domain knowledge could improve performance.


\paragraph{Replace} \textbf{Replace} $\rightarrow$ \textbf{architecture} is one of the most common architecture-based speedup approaches in the literature we reviewed.

One theme of \textbf{replace} $\rightarrow$ \textbf{architecture} work is frequency-based reformulations of layers. Computational complexity analysis suggests that replacing dense layers with Fourier-transform-based versions can produce speedups \citep{yang2015deep,moczulski2015acdc}, but \citet{yang2015deep} note that practical realization of speedups would require a large fraction of layers to be dense layers. This condition is met in modern transformer networks, and \citet{lee2021fnet} show that their FNets, which replace transformer attention layers with a 2D discrete Fourier transform (DFT), have lower training times than BERT without heavy accuracy loss. At small model sizes, FNets are on the speed-accuracy Pareto frontier, while at larger model sizes BERTs and FNet-transformer hybrids (using self-attention only in the final two layers) are preferable \citep{lee2021fnet}. Similarly, frequency-related reformulations of convolutional layers have also led to demonstrable speedups \citep{chen2019drop, dziedzic2019band}. 
Note that FNets are closely related to transformer variants that aim to reduce the quadratic (in sequence length) complexity of attention through approximations of attention.\footnote{We recommend \citet{tay2020efficient} for a survey on efficient transformers and \citet{transformerCodebases} for an analysis of various transformer replacement approaches.} However, FNets are notable in that they do not seek to approximate attention and instead simply mix tokens in the sequence and hidden spaces through a 2D DFT. Other efficient transformer variants use attention but avoid the complexity associated with computing its full set of activations and are thus discussed in the relevant sections of \ref{sec:derived}.


Another \textbf{replace} $\rightarrow$ \textbf{architecture} approach is using training-aware neural architecture search (NAS). \citet{tan2021efficientnetv2} developed EfficientNetV2, a convolutional neural network optimized for
faster training rather than lower FLOPs. Their architecture search shows that the replacement of MBConv layers with fused MBConv layers in early architecture stages offers faster training. Further, a smaller expansion ratio for the MBConv layers and smaller kernel sizes with more layers were also shown to help. Interestingly, \citet{tan2021efficientnetv2} train with various image sizes to attain additional speedups. They also obtain greater accuracy by reducing model regularization strength when the image sizes are smaller, which we discuss further in Section \ref{sec:obj}.

\citet{timmons2020approximating} used function approximation techniques to develop faster replacements for the sigmoid and hyperbolic tangent activation functions. They show 10-50\% reductions in training time without affecting model quality on CPUs, noting that future work may build on their techniques to speed up training on accelerators.

Our review found many approaches to \textbf{replace} $\rightarrow$ \textbf{architecture} that use static, domain knowledge mechanisms. EfficientNetV2 \citep{tan2021efficientnetv2}, however, uses a learning-based mechanism in its explicit optimization of training speed with neural architecture search (NAS). Importantly, NAS has emerged as a popular technique in the efficiency literature \citep{ren2021comprehensive}---it can be tailored to discover fast architectures for specific compute platforms \citep{liberis2021munas} and training-accelerating enhancements for known architectures \citep{so2021primer}. Notably, since it typically involves training, NAS itself can be accelerated through efficient-training techniques like data pruning \citep{prasad2022speeding}, which we discuss more in Section \ref{sec:traindata}. See \citet{white2023neural} for a recent survey of the NAS literature.

\paragraph{Retrofit} 
\textbf{Retrofit} $\rightarrow$ \textbf{architecture} aims to reduce the $n_{\text{iteration}}$ required to reach the desired accuracy at the cost of increased $T_i$. 

For example, \citet{so2021primer} show large speedups by retrofitting various transformer backbones
with operations discovered by NAS. In particular, they introduce a depth-wise convolution after each query (Q), key (K), and value (V) projection.
They also replace the softmax with a squared ReLU activation.
Importantly, these speedups are shown to be present on various AI accelerators (GPUs and TPUs) and in various models. This retrofit approach increases $T_i$ but reduces the $n_{\text{iteration}}$ needed to reach a given accuracy. 

Another popular approach that involves a similar tradeoff is increasing layer width or number of layers (depth), which can lead to a higher accuracy gain per training step. \citet{li2020train} use this strategy and show that increasing the depth and width of RoBERTa models can produce large speedups.

Our review suggests that \textbf{retrofit} $\rightarrow$ \textbf{architecture} approaches tend to be static in nature; i.e., changes are made only once at initialization. However, we found changes may either be based on domain knowledge or learned via evolutionary search over various layer types \citep{so2021primer}. 

\subsection{Data Speedup Strategies}

Data-related speedup methods apply the 5Rs to the training data and derived data subcomponents.
Approaches in the \textbf{remove} $\rightarrow$ \textbf{derived data} category (e.g., Selective-Backprop) reduce the burden of gradient and activation computations and thus can reduce $T_i$.
On the other hand, if data loading is the bottleneck, \textbf{remove} $\rightarrow$ \textbf{training data} approaches can speed up training through methods that include reducing the sizes of data samples.
Regardless of bottleneck location, speedups can be achieved by reducing the $n_{\text{iteration}}$ required to reach the desired accuracy through methods like \textbf{reorder} $\rightarrow$ \textbf{training data} (e.g., curriculum learning). 
We place various data-related speedup strategies into our taxonomy and review them below.

\subsubsection{Training Data}
\label{sec:traindata}

\paragraph{Remove}

\textbf{Remove} $\rightarrow$ \textbf{training data} can occur at both the dataset and data sample level. On a sample level, Cutout~\citep{devries2017improved} drops square regions of pixels, and ColOut~\citep{mosaic2021colout} randomly drops rows and columns of an input image for a vision model. If the dropout rate is not too large, the image content is not significantly affected, but the image size is reduced. Both of these methods act as a form of regularization, but only ColOut reduces computation too. 

On a dataset level, several data-pruning methods are motivated by the observation that there are significant redundancies in popular benchmark datasets \citep{birodkar2019semantic,paul2021deep}. The challenge for data pruning research is to identify samples to be excluded from training without compromising model quality. 
\citet{mirzasoleiman2020coresets} propose Coresets for Accelerating Incremental Gradient descent (CRAIG). CRAIG formulates data pruning as a weighted subset selection problem that aims to approximate the full gradient using selected data. To efficiently solve this NP-hard problem, they transform the original problem into a submodular set cover problem that can be solved using a greedy algorithm.
\citet{killamsetty2021grad} propose GradMatch, which is similar to CRAIG but tries to directly minimize the gradient difference between the subset and the entire training set. The authors show that the objective function is approximately submodular and use orthogonal matching pursuit (OMP) to solve the gradient matching problem.
\citet{killamsetty2021glister} introduce GLISTER (``GeneraLIzation based data Subset selecTion for Efficient and Robust learning''). GLISTER solves a mixed discrete continuous bi-level optimization problem to select a subset of the training data by maximizing the log-likelihood on a held-out validation set. The authors use a Taylor-series approximation to make GLISTER efficient. All of these approaches make use of the connection between submodularity and data selection that was explored in earlier work on dataset reduction for speech recognition \citep{wei2013using, wei2014unsupervised}, machine translation \citep{kirchhoff2014submodularity}, and computer vision \citep{wei2015submodularity}. Further, they have been built upon to accelerate domain adaptation training \citep{karanam2022orient}, semi-supervised learning \citep{killamsetty2021retrieve}, and hyperparameter tuning \citep{killamsetty2022automata}.

Relatedly, \citet{raju2021accelerating} speed up training by determining a new subset of data to train on multiple times throughout training via simple heuristics for calculating sample importance. They attribute their data pruning approach's ability to maintain model quality at aggressive pruning rates to its dynamism and the existence of points that are important to the learned decision boundary for only some of the training time. To better exploit these points, they propose two reinforcement-learning-based sample importance scores, use them for dynamic data pruning, and achieve slightly better accuracy than a random (but faster) baseline.

The aforementioned approaches perform data selection every $L \geq 1$ epochs of stochastic gradient descent, and the gradient descent updates are performed on the subsets obtained by the data selection. However, a number of works have also studied the problem of pruning a dataset prior to (or very early in) training. For example, \citet{toneva2018empirical} show that a large fraction of data can be removed prior to training without accuracy loss by removing the examples associated with the lowest ``forgetting event'' counts. A forgetting event is counted when a sample transitions from being classified correctly to misclassified during training, as computed by training the model on the full dataset. Due to the need to train the model on the full dataset before training it on the forgetting-event-based subset, they experiment with computing forgetting events on a faster-to-train model; they show that this leads to effective subsets for the original model.
Consistent with this, \citet{coleman2019selection} design the ``selection via proxy (SVP)'' approach that performs subset selection based on properties computed with a faster-to-train proxy model. These proxy models are obtained by removing hidden layers from the target deep model, using smaller architectures, and training for fewer epochs. Despite having lower accuracy, these models can be used to compute characteristics like forgetting events that can be used to select data subsets that are effective for training a high-quality target model more quickly. SVP was able to prune 50\% of CIFAR-10 without harming the final accuracy. 
\citet{chitta2021training} show that data pruning can be performed by leveraging ensemble active learning. In particular, they build implicit ensembles from hundreds of training checkpoints across different experimental runs, then build a data subset using prediction-uncertainty-based data acquisition functions that are parameterized by the ensembles.  
\citet{sorscher2022beyond} evaluate many existing data pruning metrics, finding that 1) they perform poorly on ImageNet, even if they performed well on smaller datasets like CIFAR-10; 2) the best metrics are computationally intensive; and 3) all depend on labels (which precludes data pruning for unlabeled dataset pretraining). They address these disadvantages by ranking example difficulty without label access via k-means clustering in the embedding space of an ImageNet model trained with self-supervision. Examples with embeddings far from the nearest cluster centroid are considered the most difficult and kept in the dataset. This approach allows training on ImageNet with 20\% fewer data samples, correspondingly lower $n_{\text{iteration}}$, and little accuracy loss. Interestingly, \citet{sorscher2022beyond} also provide evidence for the idea that the hardest (easiest) examples should be kept when the initial training set is relatively large (small).

As our discussion reflects, the \textbf{remove} $\rightarrow$ \textbf{training data} approaches we found span both static and dynamic mechanisms. While learning-based mechanisms typically produce the best results, cheaper-to-compute domain-knowledge-based mechanisms like random pruning can be effective, particularly when applied dynamically. Significant progress has been made on pruning smaller datasets like CIFAR-10 and CIFAR-100, but we are still far from achieving large amounts of data pruning on ImageNet-scale data. 

\paragraph{Restrict} 
\textbf{Restrict} $\rightarrow$ \textbf{training data} can be done at both the sample and dataset levels. These techniques tend to achieve speedup by reducing $n_{iteration}$ while keeping $T_i$ fixed. The major idea behind this class of speedup techniques is to restrict the statistical properties of the data.

On a data sample level, popular normalization techniques include centering, scaling, decorrelating, standardizing, and whitening.
A more detailed review of these techniques is provided in \citet{huang2020normalization}.


On a dataset level, there are several theoretical works that show that significant speedups can be achieved by imposing certain restrictions on the training data generation process. For example, \citet{kailkhura2020statistical} show that training datasets with optimized spectral properties, e.g., space-filling designs, produce models with superior generalization. \citet{jin2020quantifying} introduce cover complexity (CC) to measure the difficulty of learning a dataset as a function of the richness of the whole training set and the degree of separation between different labeled subsets. They found that the error increases linearly with the cover complexity both in theory and on MNIST and CIFAR-10. The major obstacle preventing the use of these theoretical findings is that one rarely has control over the data-generating distribution.

Most of the above \textbf{restrict} $\rightarrow$ \textbf{training data} approaches are static and based on domain knowledge. 

\paragraph{Reorder} 
\textbf{Reorder} $\rightarrow$ \textbf{training data} changes the order in which examples or aspects of examples are presented to the model during training. The central challenge of this set of approaches is outperforming the random default ordering typically used in training. We discuss ordered learning via curriculum and progression approaches and suggest \citet{wang2021survey} for further discussion of ordered learning.


Reordering the dataset using a curriculum can aid learning \citep{bengio2009curriculum}, allowing a larger performance gain per iteration that reduces the $n_{\text{iteration}}$ required to reach a desired performance level. 
\citet{hacohen2019power} analyzed the effect of key curriculum learning method components: 1) the scoring function used to assign training examples a difficulty level; 2) the pacing function used to determine when new examples are presented to the model; and 3) the ordering that determines whether examples are presented easiest first, hardest first, or randomly. Scoring functions typically reflect example complexity, uniqueness, signal-to-noise ratio, etc., while pacing functions can be linear, step-based, exponential, etc. Interestingly, \citet{hacohen2019power} found that curriculum learning offers a relatively small but statistically significant training speed benefit.
Similarly, \citet{wu2020curricula} systematically study the effect of different scoring, pacing, and ordering choices. On CIFAR-10, CIFAR-100, and FOOD-101(N), they found that scoring and ordering had no effect on model quality at convergence, while the dynamic dataset size induced by the pacing function offers a small benefit. However, \citet{wu2020curricula} found that speedup benefits (through fewer training steps required to reach a particular accuracy) were notable when there was a limited training iteration budget or when labels were noisy; in both cases, an easy-then-hard ordering was helpful, while other orderings were not.

Ordered learning of data can also be done in a progressive manner, with an important example being progressive resizing \citep{howard2018resizing,touvron2019fixing,hoffer2019mix, tan2021efficientnetv2}. This method works by initially training on images that have been downsampled to a smaller size, then slowly growing the images to a larger size. Training is accelerated because computation in many models is proportional to image resolution, so $T_i$ is reduced for most of the training. Accuracy is maintained (or even improved) with these approaches due to the model's exposure to various image sizes during training. Interestingly, \citet{tan2021efficientnetv2} find that the performance of this approach is further improved by increasing the regularization strength for higher image resolutions. 

Our review found that \textbf{reorder} $\rightarrow$ \textbf{training data} is mostly approached with domain knowledge mechanisms. An exception is ``teaching with commentaries'' \citep{raghu2020teaching}, which uses a learning-based mechanism to find meta-information (or commentaries) that will produce speedups when used during training. Specifically, \citet{raghu2020teaching} show that learned
per-iteration example weights can facilitate learning speed improvements. 
More work along these lines may help the success of \textbf{reorder} $\rightarrow$ \textbf{training data} approaches, as the marginal benefits seen with domain-knowledge-based curricula may relate to not-yet-understood aspects of neural network training dynamics.

\paragraph{Replace} There are two broad approaches to \textbf{replace} $\rightarrow$ \textbf{training data} in the literature we reviewed: 1) replacing the training dataset with an encoded/compressed version; and 2) replacing the training dataset with a distilled dataset. Both sets of approaches usually reduce training step time $T_i$ by operating on smaller and/or more efficient representations. The former approaches typically require changes to the model but can avoid the expensive JPEG decoding process. For instance, \citet{gueguen2018faster} train directly on DCT coefficients, avoiding the JPEG decoding/decompression step. Beyond requiring model changes, a limitation of this method is that usage of traditional data augmentation strategies will require a decoding-augmentation-reencoding step, which may eliminate the speed gains. \citet{dubois2021lossy} learn an encoder that can perform zero-shot compression of new training datasets, achieving not only substantial bit-rate savings relative to JPEG but also encodings that can be learned from directly. Learning a compressor that aims to preserve predictability (rather than human interpretability) is an apparently new but promising direction: \citet{dubois2021lossy} provides a minimal script that trains an image encoder, encodes the STL dataset, and trains a linear classifier on the resulting encodings to 98.7\% accuracy in under five minutes.

Another way in which data can be replaced is through data distillation \citep{wang2018dataset, zhao2020dataset, nguyen2020dataset, nguyen2021dataset}. These approaches replace the full set of training samples with a smaller set. Unlike subset selection, these approaches produce samples that may not appear in the original dataset, and can instead be the output of an arbitrary algorithm.
Dataset distillation is often expressed as a bi-level meta-learning process where an ``inner loop'' trains a model on a distilled dataset, and an ``outer loop'' learns to compress that data for performance on the original dataset. Compared to training on a non-distilled dataset of equivalent size, higher performance per step of training can be reached when using distilled data. However, our literature review suggested that data distillation approaches may be less helpful for reaching the performance levels attainable by training with a large non-distilled dataset.

Mechanistically, the \textbf{replace} $\rightarrow$ \textbf{training data} approaches we found are primarily static, learning-based methods. Reduced model quality is a major concern when replacing the data, and it's possible that applying \textbf{replace} $\rightarrow$ \textbf{training data} more dynamically (e.g., at different points during training) might offer more control over the speedup-quality tradeoff.

\paragraph{Retrofit} \textbf{Retrofit} $\rightarrow$ \textbf{training data} is approached in many ways in the literature. These techniques mostly aim to improve 
$n_{\text{iteration}}$.

Data augmentation is a widely used method for retrofitting a dataset with more examples---see \citet{shorten2019survey} and \citet{feng2021survey} for image and NLP surveys. For example, \citet{dai2021coatnet, wightman2021resnet} show that RandAugment \citep{cubuk2020randaugment}, mixup \citep{zhang2017mixup}, and CutMix \citep{yun2019cutmix} can improve accuracy.
Further, \citet{hoffer2019augment, fort2021drawing, wightman2021resnet} accelerate data loading (and sometimes increase accuracy) by using multiple augmentations of a given image in each batch.
\citet{fort2021drawing} provides evidence that this approach speeds up learning by reducing gradient variance arising from the data augmentation process---for a fixed batch size, the reduction in variance achieved by raising the augmentation multiplicity enables stable training with a larger learning rate and higher performance per step. See \citet{fort2021drawing} to learn more about the interaction between augmentation multiplicity, learning rate, and performance/speedup.
\citet{choi2019faster} use the same examples (e.g., the same batch) for multiple training iterations in an epoch when the training process is CPU bound, which accelerates training by ensuring the AI accelerator doesn't idle. 

Orthogonally, pre-training on large datasets that may have no or weak labels is an essential (though expensive) retrofit strategy for attaining the highest possible accuracy \citep{devlin2018bert, dai2021coatnet, zhai2021scaling}. Indeed, \citet{hoffmann2022training} show that scaling up the dataset size and model size roughly equally yields the best quality for a fixed compute budget---at least for language modeling with transformers.

Our review found that \textbf{retrofit} $\rightarrow$ \textbf{training data} is usually approached with a domain knowledge mechanism. One exception is \citet{raghu2020teaching}, where the authors learned augmentations to improve generalization. Learning-based \textbf{retrofit} $\rightarrow$ \textbf{training data} methods like AutoAugment \citep{cubuk2018autoaugment} are difficult to develop due to the computational demands of learning augmentation policies on large-scale tasks, combined with the difficulty of transferring policies learned on other (smaller) tasks \citep{cubuk2020randaugment}.

\subsubsection{Derived Data}
\label{sec:derived}
Our survey found that speedups obtained by targeting derived data (i.e., activations and gradients) primarily involve the \textbf{remove} and \textbf{restrict} actions.
Since the targets of the speedup method are created during training (e.g., activations don't exist until the data is fed through the model), these approaches necessarily are applied during the training procedure rather than before training.
Their behavior can be learning-based (e.g., learnable attention masks) or based upon domain knowledge (e.g., removing high-frequency components of feature maps). 

\paragraph{Remove}
\textbf{Remove} $\rightarrow$ \textbf{derived data} aims to reduce the time per training step $T_i$ by removing activation or gradient information. A key challenge is developing a good approximation to or proxy for a training step that uses all activation and gradient information. 

For convolutional layers, redundant or less helpful feature map components can be removed  using frequency information \citep{chen2019drop, dziedzic2019band}. \citet{chen2019drop} downsample a fraction $\alpha$ of a feature tensor's channels by a factor of 2 in each spatial dimension and modify the filters to operate on these lower resolution channels, achieving lower complexity for the computations involving the downsampled portion of feature maps---the parameter $\alpha$ controls the tradeoff between accuracy and speedup for these ``Octave Convolutions'' that replace regular convolutions. Relatedly, \citet{dziedzic2019band} introduce band-limited convolutional layers, retaining the original feature map size in the spatial domain but performing convolution in the frequency domain via FFT, which allows the computational exclusion of feature map components according to their frequencies. \citet{dziedzic2019band} choose to exclude higher frequencies from the FFT based on the idea that models are biased towards lower frequencies. They show that the resulting models are accurate, with the fraction of frequencies discarded governing the accuracy-speedup tradeoff.

Similarly, there are various techniques that address the well-known quadratic (in sequence length) memory and time complexity of the attention operation by removing entries from the attention matrix. This is done in either a fixed manner using the domain knowledge that nearby tokens are important
or in a learnable/dynamic manner.
Examples of the former include the Memory Compressed Transformer \citep{liu2018generating} and Image Transformer \citep{parmar2018image}, while the Sinkhorn Transformer\citep{tay2020sparse} illustrates the latter. 
Because the central idea of these methods is to remove the quadratic complexity with respect to sequence length, they are most beneficial in situations that involve long sequences. Accordingly, \citet{tay2020long} introduce a suite of long-range tasks to compare the various ``efficient'' attention mechanisms. \citet{tay2020long} suggests that dynamic/learnable approaches are more accurate but slower than static/domain-knowledge-based approaches (see Figure 3 of \citeauthor{tay2020long}, \citeyear{tay2020long}), consistent with what our survey tended to find for dynamic/learnable versions of other speedup approaches.
\citet{tay2020efficient} provides a survey of methods for improving attention's efficiency, and we describe several more of these methods below in terms of other 5R speedup actions being applied to the derived data (i.e., the full attention matrix activation set, which has quadratic complexity).

Since the backward pass can be twice as expensive as the forward pass, significant speedups can be obtained by only computing gradients for a subset of examples that meet some forward-pass-related criteria. Motivated by importance sampling \citep{loshchilov2015online, katharopoulos2018not}, \citet{jiang2019accelerating} propose Selective-Backprop, a simple but effective gradient pruning technique. Given the loss for a sample, Selective-Backprop chooses to either include or omit this sample from the backward pass.
Experiments on CIFAR-10, CIFAR-100, and SVHN show that Selective-Backprop can achieve a 3.5x speedup compared to standard SGD in exchange for a decrease in accuracy.
\citet{mindermann2021prioritized} similarly compute gradients for only a subset of samples but select them based on more than simply the magnitude of their loss. Indeed, they find that high-loss samples can slow down training when they correspond to noisy labels or rare exceptions. Therefore, they propose Reducible Holdout Loss Selection (RHO-LOSS) to identify the examples that are expected to improve generalization. Specifically, they only compute gradients for samples that produce a high loss for the model being trained but a low loss for a small proxy model trained on a held-out dataset. Similarly, \citet{siddiqui2022metadata} accelerate training by only computing gradients for samples with both high loss and high probability of being clean (e.g., not mislabeled) according to their Metadata Archaeology via Probe Dynamics (MAP-D) sample-classification approach.

Our review found that \textbf{remove} $\rightarrow$ \textbf{derived data} is mostly approached using dynamic, domain-knowledge-based mechanisms. Interestingly, the types of domain knowledge used in the methods we reviewed suggest that making approximations to learning-based speedup approaches is a fruitful path towards new \textbf{remove} $\rightarrow$ \textbf{derived data} speedup methods. For example, \citet{parmar2018image,liu2018generating} use fixed sparse attention that is an approximation to learnable sparse attention \citep{tay2020sparse}, and \citet{mindermann2021prioritized} develop approximations to avoid solving an intractable holdout-loss minimization problem over the set of possible points to train on.

\paragraph{Restrict} Our review found that \textbf{restrict} $\rightarrow$ \textbf{derived data} is applied during both the forward and backward passes. Like weights, gradients and activations can be restricted via quantization or low-rank approximations to reduce $T_i$. Also, normalizing or otherwise restricting the distribution of derived data can avoid ill-conditioned loss landscapes and training instability \citep{huang2020normalization}, accelerating training by reducing $n_{\text{iteration}}$. However, these methods must be careful to not introduce too much overhead or training instability via approximations; for example, the large time cost of standardizing derived data with Batch Normalization is shown in Figure \ref{fig:rn50_microbench}.

Training efficiency can be improved by normalizing activations using estimates of their statistics \citep{wiesler2014mean,desjardins2015natural} or using exact statistics computed inside normalization layers \citep{ioffe2015batch,ba2016layer,wu2018group}. 
For example, \citet{wiesler2014mean} introduce mean-normalized SGD, a preconditioning approach to keeping activations mean-centered that improves model quality and convergence rate. 
Alternatively, Batch Normalization \citep{ioffe2015batch} layers standardize activations using mini-batch statistics, ensuring activations seen during training always have zero mean and unit variance. To avoid issues with computing statistics in a way that depends on batch size, \citet{ba2016layer} introduce layer normalization, and \citet{wu2018group} generalize layer normalization with group normalization. For more discussion of activation normalization methods, please see \citet{huang2020normalization}.

Approaches that normalize gradients also exist and often aim to exploit curvature information to improve the optimization trajectory.
\citet{pascanu2013difficulty} propose clipping the gradient's norm to a threshold in order to stabilize training and improve the model quality attained per iteration. Similarly, \citet{yu2017block} propose normalizing gradients in a layer-wise manner (block normalized gradient descent) to avoid vanishing/exploding gradients. In particular, they find better performance per step when scaling the gradients 
with respect to the weights such that their norms are proportional to the norms of their associated weights.
Other gradient normalization approaches have helped training in large batch \citep{you2017large} and mixed batch-size \citep{hoffer2019mix} contexts.
Instead of scaling, \citet{yong2020gradient} propose gradient centralization, which computes the mean of each gradient tensor along some axis
then subtracts these means
so that the resulting centralized gradients have zero mean. \citet{yong2020gradient} show that this approach accelerates training; they attribute this speedup to the stabilization brought by the reduced scale of the gradient and the regularization from ensuring that the sum of weights associated with a neuron/channel is constant during training.
For more discussion of gradient normalization methods, please see \citet{huang2020normalization}.

In distributed training contexts, the cost of communicating the full gradient can be the bottleneck; restricting gradients via compression techniques can therefore provide speedups  \citep{xu2020compressed}.
For example, \citet{seide20141} propose 1-bit SGD in which 32-bit gradient elements are approximated by 1-bit representations; this leads to significant speedups with little to no accuracy loss. The accuracy preservation is a result of their ``error feedback'' mechanism, which adds the cumulative quantization error to the gradient before quantization.
\citet{bernstein2018signsgd} also quantize gradients to 1 bit via signSGD, which approximates negative gradient components with -1 and others with 1. 1-bit Adam \citep{tang20211} and 0/1 Adam \citep{lu2022maximizing} make extreme gradient compression compatible with the Adam optimizer. \citet{dettmers20158} show that 32-bit gradient data can be approximated with a variety of 8-bit representations that do not significantly increase error but greatly reduce memory requirements for large models.
Beyond quantization, gradients can be compressed through low-rank decomposition of the gradient via methods like PowerSGD \citep{vogels2019powersgd} and GradZIP \citep{cho2019gradzip}, as well as through sparsification/remove methods. 
For a more thorough discussion of gradient compression techniques, please see \citep{xu2020compressed}.

Low-rank approaches have also been used to restrict transformer activations to achieve speedups. For instance, Linformer \citep{wang2020linformer} removes the quadratic complexity in sequence length $N$ of transformer attention to obtaining faster training on long sequences. They do this by imposing a low-rank restriction on the attention matrix, the $N\times N$ matrix of activations that multiplies the $N \times d$ value (V) matrix. Specifically, \citet{wang2020linformer} add new $k \times N$ matrices E and F to the attention operation in order to project the key (K) and value (V) matrices down from $N \times d$ to $k \times d$, turning attention from $Softmax(QK^T)V$ into $Softmax(Q(EK)^T)FV$. 

Our review found that \textbf{restrict} $\rightarrow$ \textbf{derived data} is approached in both static and dynamic ways. However, most approaches depend on domain knowledge to achieve speedups, suggesting that learning-based approaches may be an interesting target for future development. 

\paragraph{Reorder} 
Our review found few ways to \textbf{reorder} $\rightarrow$ \textbf{derived data} to produce a speedup, which may reflect the difficulty of developing approaches in this area.

Exemplifying such approaches, \citet{matani2021channels} reorder the input and activation axes into a channels-last memory format and show that this can accelerate training. This is partly a product of memory access patterns and partly an artifact of software support, but mostly a consequence of the channels-last format enabling a more efficient reduction of convolution to large, contiguous, implicit GEMM operations in accelerators.


The \textbf{reorder} $\rightarrow$ \textbf{derived data} speedup methods we found \citep{matani2021channels, dao2022flashattention} use static, domain-knowledge-based mechanisms. The scarcity of approaches in this category suggests there may be opportunities for more, potentially challenging work in this direction. 

\paragraph{Replace}
Our review found several ways to \textbf{replace} $\rightarrow$ \textbf{derived data}, including activation- and gradient-focused approaches. Most approaches focused on reducing $T_i$. A major challenge in this area is reaching the model quality attained with the original derived data.

Replacing gradients with their stale counterparts can accelerate training when gradient communication is a bottleneck.
Such asynchronous updates are particularly effective when the learning rate is reduced more for gradients that are more stale \citep{dutta2018slow}. Relatedly, \citet{jiang2019accelerating} further accelerated training with Selective-Backprop---removal of the backward pass on low loss samples---by using stale losses; this is a clear improvement when a sample's loss does not change across epochs.

Kernel-based approaches to transformer attention such as Linear Transformer \citep{katharopoulos2020transformers}, Performer \citep{choromanski2020rethinking}, and Scatterbrain \citep{chen2021scatterbrain} make transformers faster on long sequences through replacement of the traditional attention matrix $Softmax(QK^T)$ with the product of the kernel feature representations of the query (Q) and key (K) matrices: $\phi(Q)\phi(K)^T$. This allows the costly softmax attention to be approximated or replaced with a kernel formulation $\phi(Q)\phi(K)^TV$. When reordered as $\phi(Q)(\phi(K)^TV)$, this formulation has linear complexity with respect to the sequence length. 

The \textbf{replace} $\rightarrow$ \textbf{derived data} methods discussed are based on domain knowledge. Methods that alter the attention matrix, including those reviewed here, could be viewed as static replacements of the architecture. We emphasize their effect on derived activations by classifying them as dynamic modifications of the attention matrix, perhaps suggesting new efficient transformer approaches that utilize ideas from other derived-data approaches.

\paragraph{Retrofit} There exist a few ways to \textbf{retrofit} $\rightarrow$ \textbf{derived data} to achieve speedup. 

\citet{neelakantan2015adding} propose a low-overhead and easy-to-implement speedup technique: add time-dependent Gaussian noise to the gradients at every training step. The authors found that adding annealed Gaussian noise (i.e., decaying the variance over time) can improve model quality significantly, reducing the $n_{\text{iteration}}$ required to achieve the desired quality.

\citet{press2021train} propose Attention with Linear Biases (ALiBi), which adds  a penalty to attention activations that encourages attention at different length scales for different heads. This enables models trained on shorter sequence lengths, which can be trained more quickly, to perform competitively with baseline models when evaluated on longer sequence lengths at test time.

Our review found that  \textbf{retrofit} $\rightarrow$ \textbf{derived data} is mostly approached using dynamic, domain knowledge mechanisms. We expect that more accurate methods could be designed by leveraging learning-based mechanisms.

\subsection{Optimization Speedup Strategies}
\label{sec:optimization}
Optimization-related speedup methods apply the 5Rs to the optimization algorithm and training objective subcomponents. Most of them aim to minimize the number of iterations $n_{\text{iteration}}$ required to achieve the desired model quality. For example, a \textbf{retrofit} $\rightarrow$ \textbf{training objective} approach that adds sharpness information to the loss function can improve model quality gained per optimization step \citep{foret2020sharpness}. Often, these methods require increased computation. When this is the case, they are most useful when the accelerators are underutilized due to communication or data loading bottlenecks.

\subsubsection{Training objective}
\label{sec:obj}

\paragraph{Remove} 
\textbf{Remove} $\rightarrow$ \textbf{training objective} primarily aims to speed up training by reducing $T_i$ through approximation of expensive loss functions. As with many remove/restrict approaches, a key challenge here is developing an approximation that maintains the model quality associated with the original loss function.

Loss functions can be expensive to compute when neural networks have large output spaces. For example, when the vocabulary size is large, the final softmax classifier can become a bottleneck \citep{chen2015strategies}.

To combat this bottleneck, \citet{morin2005hierarchical} introduce the hierarchical softmax approach to compute the probability of a target output word as a product of probabilities associated with nodes along a path of a binary tree having leaves that span the entire vocabulary.
This requires work logarithmic in the number of classes, rather than linear.
Sampling-based approaches achieve a similar removal through different means; \citet{mikolov2013distributed} develop negative sampling, which removes from the objective all word probabilities except the target word's and those for several negative sample words drawn from a noise distribution. 

Loss functions can also be expensive to compute when they include higher-order information like loss surface sharpness. Accordingly, \citet{liu2022towards} focus on improving the efficiency of Sharpness-Aware Minimization (SAM) \citep{foret2020sharpness}, which uses perturbed weights at each iteration to bias optimization towards flatter minima. Specifically, \citet{liu2022towards} propose 
to reuse the same weight perturbation for SAM across several iterations, greatly reducing overhead while preserving accuracy.
Similarly, \citet{du2021efficient} compute the SAM weight perturbation for only a subset of weights and using only a subset of samples.
They find that these removals greatly accelerate SAM with no loss of model quality. More recently, \citet{du2022sharpness} introduced Sharpness-Aware training for Free (SAF), which removes  the weight perturbation term from the SAM objective and uses changes in the loss over time to recover an analogous sharpness-aware objective; this approach has roughly the same computational cost as vanilla training and accuracy/sharpness-avoidance benefits similar to SAM's, which can speed up training by reducing the number of iterations required.

The \textbf{remove} $\rightarrow$ \textbf{training objective} strategies we reviewed depended on domain-knowledge based mechanisms that offered limited control over the drop in model quality caused by the removal. In future work, it may be fruitful to dynamically choose the fidelity of one's approximation to SAM, softmax, etc.

\paragraph{Restrict} 

The only \textbf{Restrict} $\rightarrow$ \textbf{training objective} method we found is that of \citet{akbari2021does}, who restrict the geometry of the loss landscape by altering the loss function. 
Specifically, they introduce the Generalized Jeffries-Matusita loss, which has a smaller Lipschitz constant than a Kullback-Leibler divergence loss. They show that their loss yields lower generalization error after a fixed number of steps. It also enables tighter generalization guarantees.


Our review suggests that \textbf{restrict} $\rightarrow$ \textbf{training objective} is still in an early stage. The approach we reviewed relies on domain knowledge, and we expect that more accurate methods could be designed by developing a deeper understanding of the interaction between loss landscape and generalization error. 

\paragraph{Reorder} 
We found two methods that \textbf{reorder} $\rightarrow$ \textbf{training objective}, both of which involve changing the loss function and/or regularization throughout training.

\citet{wu2018learning} propose Learning to Teach with Dynamic Loss Functions (L2T-DLF), in which the loss function at iteration $t$ is the output of a teacher model. This teacher model's input consists of $t$ and several variables describing the student model's performance. The authors show how to optimize the teacher model such that the series of loss functions it outputs lead to more model quality gained per student model weight update, reducing $n_{\text{iteration}}$ on both vision and language tasks.

Similarly, \citet{tan2021efficientnetv2} show that more model quality is gained per training step when regularization is adjusted as inputs are resized: they lower $n_{\text{iteration}}$ without accuracy loss by jointly increasing image resolution, dropout rate, and data-augmentation magnitude throughout training.

While these methods use both knowledge- and learning-based mechanisms, the knowledge utilized and learned was somewhat specific to the tasks considered. This may mean there are opportunities for more general \textbf{reorder} $\rightarrow$ \textbf{training objective} speedup methods. E.g., perhaps one could generalize the resolution-dependence in the image model of \cite{tan2021efficientnetv2} to sequence-length dependence in language models. There might also be opportunities to employ information about the current loss landscape or different phases of training  \citep{achille2017critical,frankle2020early}.

\paragraph{Replace} 
\textbf{Replace} $\rightarrow$ \textbf{training objective} is a common technique for achieving speedups through improvements to model quality gained per optimization step. Work in this area is typically faced with the challenge of outperforming common objectives like cross-entropy or mean squared error (MSE).

When using mixup and CutMix data augmentation, \citet{wightman2021resnet} replace the typical cross-entropy loss with one designed to align better with human perception, speeding up training of ResNet-50 on ImageNet. These specific augmentations are relevant because they produce images and labels that are combinations of multiple images (usually from different classes) and their labels; for example, mixup's images have the form  $\tilde{x} = \lambda x_i + (1-\lambda) x_j$ and mixup's labels have the form $\tilde{y} = \lambda y_i + (1-\lambda) y_j$, where $y_i,y_j$ are the one-hot labels for images $x_i,x_j$ and $\lambda \in [0,1]$. Optimizing the cross-entropy loss applied to the predicted label vector and $\tilde{y}$ as is typical encourages the model to predict a $\lambda$ probability of the presence of class $y_i$ in image $\tilde{x}$, whereas humans can typically recognize the presence of class $y_i$ in $\tilde{x}$ with certainty. Therefore, \citet{wightman2021resnet} instead assume that all classes used to construct $\tilde{x}$ are discernibly present in $\tilde{x}$ by setting each of their labels to $1$. To accommodate this change, they treat the problem as a multi-label classification and optimize the binary cross-entropy (BCE) loss for each class.

\citet{gonzalez2020improved} apply 
genetic programming to learn loss functions from primitive operations using MNIST validation dataset performance as a signal. They term their final learned loss function the ``Baikal loss.'' Despite being discovered via MNIST data, Baikal continued to provide improvements relative to cross-entropy when evaluated on CIFAR-10 data.
Along the same lines, \citet{bechtle2021meta} propose a meta-learning approach to training parameterized loss functions, Meta-Learning via Learned Loss (ML$^3$). On previously unseen tasks, the authors show that the meta-learned loss functions improve convergence speed.

In terms of mechanisms, our review found that \textbf{replace} $\rightarrow$ \textbf{training objective} is approached mostly in a static manner. 
We suspect that dynamic mechanisms---i.e. altering the loss function used based on the data and state of training---may facilitate further improvements in the emerging learning-based \textbf{replace} $\rightarrow$ \textbf{training objective} literature \citep{bechtle2021meta, gonzalez2020improved}.

\paragraph{Retrofit} 
\textbf{Retrofit} $\rightarrow$ \textbf{training objective} is a straightforward way to 
reduce $n_{\text{iteration}}$.
This popular set of approaches features a variety of ways to incorporate domain-specific information or known learning principles as additional objectives.

\citet{von2019informed} survey various ways to inject prior knowledge into loss functions to accelerate training and improve model quality. They present approaches in terms of how the knowledge is represented (e.g., as algebraic equations) and its source (e.g., physical laws). Generally, these approaches accelerate training by penalizing model inconsistencies with user-provided knowledge about the data.
\citet{wang2021physics} survey methods for incorporating prior knowledge of physics, many of which entail adding terms to the loss.
\citet{kukavcka2017regularization} discuss various application-agnostic ways to regularize deep learning. For example, adding an $L_2$ penalty is a common and often helpful practice.

Perhaps the most notable recent work that retrofits the training objective is sharpness aware minimization (SAM) \citep{foret2020sharpness}. Based on the observation that flatter loss minima are associated with better generalization \citep{keskar2016large}, \citet{foret2020sharpness} add sharpness information to the training objective to encourage convergence to a flatter minimum. SAM produces better model quality per step, but its steps require much more time to execute due to the use of an additional gradient computation to determine sharpness. 
However, efficient variants of SAM exist, which we discuss in the \textbf{remove} $\rightarrow$ \textbf{training objective} section.

Regarding mechanisms, the \textbf{retrofit} $\rightarrow$ \textbf{training objective} methods we reviewed rely heavily on domain knowledge. This suggests that deepening our understanding of learning principles might provide means to further enhance training objectives. There may also be opportunities for speedups through methods that learn a new element of the training objective rather than learning the entire loss function \citep{bechtle2021meta}.

\subsubsection{Algorithm} \label{sec:optAlgo}

\paragraph{Remove} 
The high-level algorithm for gradient-based training of neural networks (stochastic gradient descent, or SGD) is extraordinarily simple given its effectiveness. Accordingly, a major challenge for approaches that seek to \textbf{remove} $\rightarrow$ \textbf{algorithm} is preservation of model quality.

Removal of all gradient computations is one path toward accelerating training. Indeed, there are various zeroth-order optimization (ZOO) training algorithms (see \citeauthor{liu2020primer}, \citeyear{liu2020primer}, for a review). 
For example, Zeroth-Order Relaxed Backpropagation (ZORB) \citep{ranganathan2020zorb} starts its backward pass with the true sample labels at the output layer, then successively applies the pseudoinverse operation to each layer's weights
to determine what each preceding layer should output in order to predict the labels at the output layer. A single forward pass then applies the pseudoinverse to the data sample input to solve for the first-layer weights that will produce the desired first-layer output determined by the backward pass. The output of this layer then becomes the input used to solve for the next layer's weights, and so on. Using small networks on small datasets, this approach can match the accuracy of gradient-based algorithms like Adam while being 300 times faster. However, this pseudoinverse-based approach is difficult to scale up. 

Rather than remove computationally intense steps from each iteration of an algorithm, many methods remove entire iterations through weight prediction.
For example, \citet{dogra2020optimizing} introduce Koopman training, which uses Koopman operator theory to learn and quickly predict optimization trajectories. This allows them to extrapolate future weight updates without gradient evaluations. Notably, though, Koopman training can lead to model accuracy degradation and higher memory costs due to the need to track weight evolution. \citet{sinha2017introspection} provide a similar approach but use an ``introspection network'' trained to predict weight changes on an easy task (MNIST) to predict weight changes on general tasks (e.g., ImageNet). \citet{knyazev2021parameter} remove all optimizer iterations and simply predict performant parameters using a Graph HyperNetwork (GHN) in under one second. While this prediction network can be applied to unseen architectures, it must be trained initially 
and its predicted weights are not as performant as weights optimized by gradient-based methods. 

\textbf{Remove} $\rightarrow$ \textbf{algorithm} mostly relies on computationally-intense, learning-based mechanisms. However, consistent with the difficulty of removing elements from an already simple training algorithm, these methods often show significant accuracy degradation, especially on moderate to large-scale data such as ImageNet. We nonetheless expect more work in this emerging area
as neural network training becomes better understood.

\paragraph{Restrict}  
The \textbf{restrict} $\rightarrow$ \textbf{algorithm} approaches we review reduce $n_{\text{iteration}}$ and thereby produce speedups by ensuring that training stays in an efficient regime (e.g., avoiding exploding or vanishing gradients). Mostly based on domain knowledge, these approaches depend on insights regarding choices like initialization scale and step size.

For example, \citet{he2015delving} restrict the initialization of weights in deep DNNs with $\text{ReLU}$ activations to a symmetric distribution with variance $2/\text{fan-in}$.  \citet{hanin2018start} provide a theoretical framework for understanding the importance of this restriction, showing that either uniform or normal weight-initialization distributions that employ it avoid failure modes that can arise with other distributions.

\citet{vorontsov2017orthogonality} restrict the weights at initialization and during optimization to be approximately orthogonal via a spectral margin parameter that controls deviations from orthogonality, with a margin of zero corresponding to exact orthogonality. They find that baseline RNNs without any margin constraint can converge slowly or not at all, while networks trained with margins slightly greater than 0 converged faster than both unconstrained and 0-margin networks. This result suggests a benefit to balancing between a hard constraint and no constraint at all.
\citet{hu2020provable} provide theoretical evidence supporting the idea that orthogonality restrictions simply at weight initialization (rather than throughout training) can be sufficient to speed up convergence.

\citet{mccandlish2018empirical} theoretically and empirically support the idea that the optimal batch size depends on the interaction between the batch size and the ``gradient noise scale,'' where the latter measures the inter-batch variability of the gradients. They find that when the noise scale is large relative to the batch size, one can increase the batch size to decrease the number of optimization steps required, providing a speedup when parallelism prevents an offsetting increase in $T_i$. 
\citet{smith2017don} find that one can also increase the batch size by using this as a substitute for decreasing the learning rate.

When using large learning rates, \citet{smith2017super} find that 
restricting the level of weight decay to account for the current learning rate can help avoid excessive regularization.

The \textbf{restrict} $\rightarrow$ \textbf{algorithm} mechanisms in our review were mostly static and domain-knowledge based. We suspect that more dynamic \textbf{restrict} $\rightarrow$ \textbf{algorithm} mechanisms based on domain knowledge could produce further speedups. For example, the analyses of \citet{mccandlish2018empirical} and \citet{wang2022provable} suggest that the batch size and momentum parameters (respectively) should be adaptively set due to their dependence on dynamic quantities.

\paragraph{Reorder}
\textbf{Reorder} $\rightarrow$ \textbf{algorithm} is mostly done to alter the progression through a learning rate (LR) schedule. Providing an effective schedule for a given task and model is the main challenge in this area.

When adjusting $n_{\text{iteration}}$ while using a cosine annealing schedule, \citet{hoffmann2022training} find that the schedule length must be adjusted such that at least 80\% of the schedule is completed during the reduced training time window, or else accuracy suffers. Moreover, the best performance is attained when the end of the schedule coincides with the end of training. 
One can also repeat a schedule multiple times, possibly with variations each time. Such cyclic learning rate schedules have empirically and theoretically produced speedups, at least in some circumstances \citep{smith2017cyclical,loshchilov2016sgdr, oymak2021provable, goujaud2021super}.

The LR scheduling approaches we found are primarily fueled by domain knowledge. However, learning-based solutions for \textbf{reorder} $\rightarrow$ \textbf{algorithm} are emerging. E.g., \citet{baik2020meta} accelerate MAML by using a task-and-iteration-specific learning rate produced by a meta-learned hyperparameter network.

\paragraph{Replace}
\textbf{Replace} $\rightarrow$ \textbf{algorithm} covers some of the most well-known ways to speed up training. These methods generally aim to replace the optimization algorithm with one that reduces $n_{\text{iteration}}$. Fair evaluations are a major concern in this area due to interactions between optimizers and hyperparameters like weight decay that can create misleading performance comparisons \citep{schmidt2021descending}.


Perhaps the oldest work in this category is that of \citet{polyak1964some}, who introduces momentum to the gradient descent algorithm. Later, \citet{hanson1988comparing} found that adding a decay step to weight updates can improve the performance gained per iteration at negligible compute cost. This decay entails multiplying the current weights by a small constant as part of each update.

More recently, there have been many ``adaptive gradient'' methods that aim to speed up training by scaling the effective learning rate intelligently throughout training. This scaled learning rate is often specific to a given parameter and is a function of past gradients. Some of the most well-known  methods of this type include RMSProp \citep{rmsprop}, AdaGrad \citep{duchi2011adagrad}, and Adam \citep{kingma2014adam}.  


While replacing SGD with adaptive gradient methods can speed up achievement of a given performance level (e.g., see \citeauthor{dosovitskiy2020image}, \citeyear{dosovitskiy2020image}), combining them with weight decay involves subtleties. In particular, \citet{loshchilov2017decoupled} note that weight decay should be differentiated from adding an $L_2$ penalty to the loss, as the latter can produce undesired effects and worse performance.
\citet{loshchilov2017decoupled} address this performance issue with the AdamW optimization algorithm, which replaces $L_2$ regularization with weight decay to improve Adam's performance. AdamW is commonly used to train large language and vision models \citep{hoffmann2022training, dai2021coatnet}.

Analogous to the replacement of hand-designed features with neural network learned features, \citet{metz2019understanding} show that hand-designed optimizers can be replaced by faster-learned optimizers.
A limitation of this approach is the need to train a task-specific optimizer; moreover, the generalization of such optimizers across tasks is unclear. 

Making subtler changes, \citet{wightman2021resnet} illustrate that optimization generally requires careful tuning based on the task/architecture. Concretely, one can reduce $n_{\text{iteration}}$ by tuning the learning rate, weight decay, and other optimizer hyperparameters.

Parameter initialization at the start of learning provides another opportunity for replacement. Replacing the random initialization of weights with pretrained weights is a particularly popular approach to achieving speedups in this area \citep{han2021pre}. For example, \citet{brown2020language} train the 175 billion-parameter language model GPT-3 and demonstrate that it is flexible enough to provide near-SOTA accuracy on some downstream learning tasks without any fine-tuning. Critically, this means transfer learning (i.e., using a pretrained model's weights at initialization) may speed up training by reducing the $n_{\text{iteration}}$ required to reach a desired model quality to \textit{zero}. \citet{bommasani2021opportunities} provide a related transfer-learning report and introduce the term ``foundation models'' for these massive, flexible, pretrained models. Separately, \citet{finn2017model} introduce model-agnostic meta-learning (MAML) for learning initializations that facilitate rapid learning on new tasks.

Our review found numerous \textbf{replace} $\rightarrow$ \textbf{algorithm} approaches, using all possible mechanisms except static changes based on domain knowledge. 
This unexplored mechanism may offer an interesting direction for future work.

\paragraph{Retrofit}  
Our review found several ways to accelerate training with \textbf{retrofit} $\rightarrow$ \textbf{algorithm} techniques. Generally, a challenge for these techniques is avoiding increases in $T_i$ that overwhelm the reduction to $n_{\text{iteration}}$ they provide.

Adding second-order/curvature information to weight update rules
can significantly improve the performance gained per training step \citep{martens2015optimizing, gupta2018shampoo, anil2020scalable, goldfarb2020practical}. To address the greatly increased cost of computing these updates, \citet{anil2020scalable} approximates the large gradient-preconditioning matrices and also updates them only occasionally. These changes 
reduce the preconditioning overhead enough to yield a
large reduction in overall training time.

The \textbf{retrofit} $\rightarrow$ \textbf{algorithm} methods in our review use dynamic, domain-knowledge-based mechanisms. Given that the major challenge these approaches face is minimizing the overhead they create, future work could consider learning-based mechanisms as a way to incorporate the cost of the retrofitting (e.g., preconditioning steps) 
into decisions about whether or how much to apply it.

\
Table \ref{tab:class} summarizes our taxonomy's classification of the speedup methods we explicitly discussed and related methods.
\begin{table} \centering {\setstretch{0.5} \fontsize{5.5pt}{10.25pt}\selectfont{ \begin{widetable}{\textwidth}{p{.07\textwidth}p{.13\textwidth}p{.14\textwidth}p{.12\textwidth}p{.13\textwidth}p{.13\textwidth}p{.10\textwidth}p{.12\textwidth}} \toprule \scriptsize{Action} & \scriptsize{Model} & \scriptsize{Architecture} & \scriptsize{Data Sample} & \scriptsize{Dataset} & \scriptsize{Derived Data} & \scriptsize{Objective} & {\scriptsize Algorithm} \fontdimen2\font=0.01ex \\ \midrule \scriptsize{Remove} & \textbf{\citet{yuan2021mest}}, \citet{zhang2021efficient}, \citet{van2020deconstructing}, \citet{hoefler2021sparsity}, \citet{mocanu2018scalable}, \citet{price2021dense}, \citet{you2019drawing}, \citet{hubens2021one}, \citet{liu2021deep}, \citet{chen2021dsee}, \citet{sung2021training}, \citet{dey2019pre}, \textbf{\citet{chen2020earlybert}}, \textbf{\citet{lym2019prunetrain}}, \textbf{\citet{adelman2021faster}}, \textbf{\citet{chen2022coarsening}} & \citet{martens2021rapid}, \textbf{\citet{brock2021high}}, \textbf{\citet{huang2016deep}}, \citet{fan2019reducing} & \textbf{\citet{mosaic2021colout}}, \textbf{\citet{devries2017improved}}, \citet{liu2023patchdropout} & \textbf{\citet{coleman2019selection}}, \textbf{\citet{killamsetty2021grad}}, \textbf{\citet{raju2021accelerating}}, \textbf{\citet{sorscher2022beyond}}, \citet{kaushal2019learning}, \citet{birodkar2019semantic}, \textbf{\citet{mirzasoleiman2020coresets}}, \textbf{\citet{killamsetty2021glister}}, \textbf{\citet{chitta2021training}}, \textbf{\citet{toneva2018empirical}}, \citet{paul2021deep}, \textbf{\citet{mittal2022partitioned}}, \textbf{\citet{sivasubramanian2021training}}, \textbf{\citet{karanam2022orient}}, \textbf{\citet{killamsetty2021retrieve}}, \citet{wei2015submodularity} & \textbf{\citet{mindermann2021prioritized}}, \citet{jiang2019accelerating}, \citet{wimmer2020freezenet}, \textbf{\citet{dziedzic2019band}}, \textbf{\citet{chen2019drop}}, \citet{parmar2018image}, \textbf{\citet{liu2018generating}}, \citet{tay2020sparse}, \textbf{\citet{liu2018dynamic}}, \textbf{\citet{siddiqui2022metadata}} & \citet{liu2022towards}, \textbf{\citet{mikolov2013distributed}}, \textbf{\citet{morin2005hierarchical}}, \textbf{\citet{du2021efficient}}, \citet{chen2015strategies}, \textbf{\citet{du2022sharpness}} & \textbf{\citet{dogra2020optimizing}}, \textbf{\citet{knyazev2021parameter}}, \textbf{\citet{sinha2017introspection}}, \citet{liu2020primer}, \textbf{\citet{ranganathan2020zorb}} \\ \midrule \scriptsize{Restrict} & \textbf{\citet{chen2021pixelated}}, \citet{idelbayev2020low}, \citet{blalock2021multiplying}, \textbf{\citet{micikevicius2017mixed}}, \citet{ioannou2015training}, \textbf{\citet{sainath2013low}}, \textbf{\citet{hu2021lora}}, \textbf{\citet{dao2022monarch}}, \citet{liebenwein2021compressing}, \citet{gholami2021survey} & \textbf{\citet{press2016using}} & \citet{huang2020normalization} & \citet{kailkhura2020statistical}, \citet{jin2020quantifying} & \citet{xu2020compressed}, \textbf{\citet{wang2020linformer}}, \citet{pascanu2013difficulty}, \citet{you2017large}, \citet{huang2020normalization}, \citet{wiesler2014mean}, \textbf{\citet{desjardins2015natural}}, \textbf{\citet{ioffe2015batch}}, \textbf{\citet{ba2016layer}}, \textbf{\citet{wu2018group}}, \textbf{\citet{yong2020gradient}}, \textbf{\citet{yu2017block}}, \textbf{\citet{dettmers20158}}, \textbf{\citet{seide20141}}, \textbf{\citet{bernstein2018signsgd}}, \textbf{\citet{vogels2019powersgd}}, \textbf{\citet{cho2019gradzip}}, \textbf{\citet{lu2022maximizing}}, \textbf{\citet{tang20211}} & \textbf{\citet{akbari2021does}} & \citet{dettmers20218}, \citet{smith2017super}, \textbf{\citet{vorontsov2017orthogonality}}, \citet{wang2022provable}, \citet{hu2020provable}, \textbf{\citet{he2015delving}}, \citet{hanin2018start}, \textbf{\citet{mccandlish2018empirical}}, \textbf{\citet{smith2017don}} \\ \midrule \scriptsize{Reorder} & \textbf{\citet{he2021pipetransformer}}, \citet{brock2017freezeout}, \textbf{\citet{raghu2017svcca}} & \citet{huang2018learning}, \textbf{\citet{saharia2022photorealistic}}, \textbf{\citet{gong2019efficient}}, \textbf{\citet{li2022automated}}, \textbf{\citet{gu2020transformer}}, \textbf{\citet{chen2015net2net}} & \textbf{\citet{hoffer2019mix}}, \textbf{\citet{touvron2019fixing}}, \citet{howard2018resizing} & \citet{weinshall2018curriculum}, \citet{wu2020curricula}, \citet{raghu2020teaching}, \citet{hacohen2019power}, \textbf{\citet{bengio2009curriculum}}, \citet{wang2021survey} & \textbf{\citet{matani2021channels}}, \textbf{\citet{dao2022flashattention}} & \textbf{\citet{wu2018learning}} & \textbf{\citet{oymak2021provable}}, \textbf{\citet{smith2017cyclical}}, \citet{goujaud2021super}, \textbf{\citet{loshchilov2016sgdr}}, \textbf{\citet{baik2020meta}} \\ \midrule \scriptsize{Replace} & \citet{kusupati2021llc}, \citet{son2018clustering}, \citet{sarwar2017gabor}, \textbf{\citet{shen2020reservoir}} & \citet{han2021deep}, \textbf{\citet{chowdhury2021learning}}, \citet{ding2021repvgg}, \citet{moczulski2015acdc}, \citet{yang2015deep}, \citet{wu2018shift}, \citet{jeon2018constructing}, \citet{bello2021lambdanetworks}, \citet{liberis2021munas}, \textbf{\citet{tan2021efficientnetv2}}, \citet{srinivas2021bottleneck}, \textbf{\citet{dai2021coatnet}}, \textbf{\citet{bello2021revisiting}}, \textbf{\citet{lee2021fnet}}, \textbf{\citet{timmons2020approximating}}, \citet{tay2020efficient}, \citet{tay2020long}, \citet{transformerCodebases}, \citet{prasad2022speeding}, \citet{brown2022wide}, \textbf{\citet{geiping2022cramming}} & \citet{gueguen2018faster}, \citet{pistono2020training}, \citet{dubois2021lossy}, \citet{le2021image}, \citet{fu2016using} & \citet{sucholutsky2020less}, \citet{nguyen2021dataset}, \textbf{\citet{nguyen2020dataset}}, \citet{zhao2020dataset}, \textbf{\citet{wang2018dataset}} & \textbf{\citet{chen2021scatterbrain}}, \textbf{\citet{choromanski2020rethinking}}, \citet{katharopoulos2020transformers}, \citet{dutta2018slow} & \textbf{\citet{wightman2021resnet}}, \textbf{\citet{gonzalez2020improved}}, \textbf{\citet{bechtle2021meta}}, \textbf{\citet{steiner2021train}} & \citet{tessera2021keep}, \citet{metz2019understanding}, \citet{baydin2022gradients}, \citet{kingma2014adam}, \citet{loshchilov2017decoupled}, \textbf{\citet{brown2020language}}, \textbf{\citet{nakerst2020gradient}}, \textbf{\citet{nesterov1983method}}, \citet{rmsprop}, \citet{duchi2011adagrad}, \citet{han2021pre}, \citet{bommasani2021opportunities}, \textbf{\citet{finn2017model}}, \citet{silver2021learning}, \textbf{\citet{xie2022adan}}, \textbf{\citet{yang2021tuning}} \\ \midrule \scriptsize{Retrofit} & \textbf{\citet{fedus2021switch}}, \citet{gontijo2021no}, \textbf{\citet{kaddour2022stop}}, \textbf{\citet{yang2019swalp}}, \citet{li2022branch}, \textbf{\citet{izmailov2018averaging}} & \citet{so2021primer}, \textbf{\citet{li2020train}} & & \textbf{\citet{choi2019faster}}, \textbf{\citet{hoffer2019augment}}, \textbf{\citet{fort2021drawing}}, \citet{yun2019cutmix}, \textbf{\citet{zhang2017mixup}}, \textbf{\citet{cubuk2020randaugment}}, \citet{hoffmann2022training}, \textbf{\citet{cubuk2018autoaugment}}, \citet{devlin2018bert}, \citet{zhai2021scaling}, \citet{shorten2019survey}, \citet{feng2021survey} & \textbf{\citet{neelakantan2015adding}}, \textbf{\citet{press2021train}} & \citet{foret2020sharpness}, \citet{von2019informed}, \citet{wang2021physics}, \citet{kukavcka2017regularization} & \textbf{\citet{anil2020scalable}}, \citet{goyal2017accurate}, \textbf{\citet{polyak1964some}}, \citet{martens2015optimizing}, \citet{goldfarb2020practical}, \citet{gupta2018shampoo}, \textbf{\citet{hanson1988comparing}} \\ \bottomrule \end{widetable} }} \caption{\textbf{Surveyed Literature in Our Taxonomy.} Bold-font approaches explicitly show training speedups. Plain-font approaches target other efficiency metrics, survey a particular set of speedup methods, and/or have not yet provided clear evidence of training speedups to the best of our knowledge.} \label{tab:class} \end{table}

\subsection{Other Directions}
Beyond the algorithmic speedup approaches that we reviewed above, there are numerous compute-platform approaches to achieving faster training. Broadly, these approaches include faster implementations/kernels for common operations, better hardware, and parallelism. For instance, without changing hardware, FFCV \citep{ffcv} speeds up data preprocessing with just-in-time (JIT) compiled data augmentations and other data-pipeline optimizations. Data-pipeline optimizations tailored to distributed training, such as NoPFS, create large speedups as well \citep{dryden2021clairvoyant}.
There are numerous approaches to large-scale parallel training \citep{shazeer2017outrageously, ben2019demystifying, deepspeedMoE, deepspeed3dblog, narayanan2021efficient}, each with various tradeoffs depending on accelerator count, interconnect, accelerator memory, batch and input sizes, model structure, and more.


\section{Best Evaluation Practices}
\label{sec:evaluation}
Speedup comparison is a challenging problem, and existing papers rarely make satisfactory comparisons \citep{dehghani2021efficiency, wu2020curricula, blalock2020state}. Stemming largely from a lack of experimental standardization and poor evaluation protocols, this limitation could lead to missed opportunities at best and misleading conclusions at worst. To overcome this limitation, speedup comparisons must achieve the following desiderata: 1) the comparison must be fair; 2) the evaluation must be comprehensive, and 3) the results must be reliable. Next, we discuss some common pitfalls of existing evaluation practices and provide suggestions on how to avoid them.

\paragraph{Clarify the Goal}
Before discussing how to properly evaluate a speedup method, it is important to clarify its goal. E.g., is the method designed to accelerate training on natural image classification tasks or for all CNNs? Should it work for any model with an attention layer or only decoder models trained on English text? While wide applicability is desirable, improvements on a well-scoped class of problems are more useful than those on an ill-defined class of problems.

\paragraph{Report Experimental Setup and Hyperparameters}
Settings of hyperparameters such as the learning rate and weight decay can affect the speed-quality tradeoff created by a speedup method. Similarly, architecture variations and differences in data preprocessing can greatly impact the end results. 

To accurately attribute the performance gain of a speedup method to the proposed component-action change, it is important to hold the hyperparameters constant to the greatest extent feasible. Moreover, to facilitate future comparisons to one's proposed method, it should be clear in either the source code or the paper what hyperparameters were used. In the case that one also wishes to claim robustness to hyperparameter choices, it is also necessary to sweep the relevant hyperparameters and report the induced variation.


\paragraph{Control the Confounding Variables}

Training time is also strongly dependent on factors unrelated to algorithmic speedup actions. Such confounding factors can be hardware related (i.e., CPU/GPU/TPU/etc. aspects) and/or software related (i.e., libraries, data loading, and other code). For example, an architecture can have significantly different training speeds on different hardware~\citep{dehghani2021efficiency}. Similarly, different libraries (e.g., JAX, PyTorch, and TensorFlow) are known to yield different accuracies for the same architecture and dataset, and variations in software support for certain operations can create a large difference in the train time.

Because it is impractical to perform experiments on all possible hardware and software variations, one needs to ensure that methods being compared use identical hardware and software to the greatest extent possible \citep{transformerCodebases}. 

\paragraph{Perform a Complete Characterization}
Achieving speedup is inherently a multi-objective optimization problem that requires optimizing for two conflicting objectives simultaneously---accuracy and train time. The optimization outcome is characterized by the accuracy-efficiency Pareto frontier. This tradeoff curve can be obtained by varying hyperparameters related to a component-action combination (e.g., the degree of sparsity for model, or data, pruning). 


Accordingly, we recommend that authors report the tradeoff curve by varying hyperparameters at a sufficient granularity (e.g., using at least $3$ operating points) for a given dataset and architecture. The curve for one's proposed method should be shown alongside the curves for competing methods. Further, high-accuracy solutions should be prioritized when selecting operating points. Ideally, authors should also use at least three dataset-architecture pairs, including modern, large-scale datasets/models.



\paragraph{Use Simple, Strong Baselines as Sanity Checks}
In addition to the baseline training recipe, the most obvious baselines to evaluate against are the existing techniques resembling the method being proposed.

\begin{figure}[t]
\centering
\includegraphics[width=.67\linewidth]{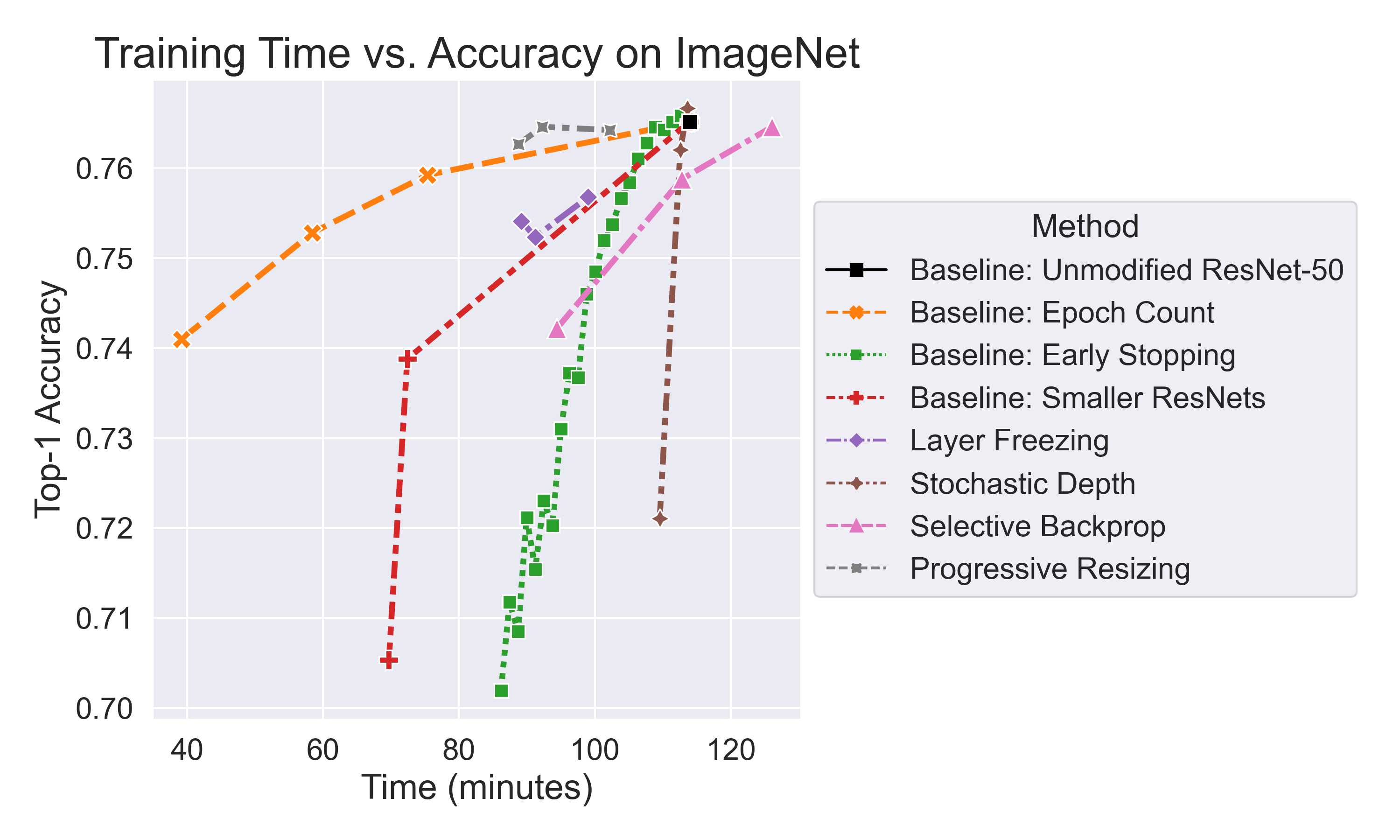}
\caption{Training for fewer epochs (``Epoch Count'') is a strong baseline, but only if one iterates through the full learning rate schedule (c.f. ``Early Stopping``). Methods that speed up training may do so at a greater cost to accuracy than simply reducing epoch count. Some methods, like Selective Backpropagation \citep{jiang2019accelerating}, may slow down training when using certain hyperparameters.}
\label{fig:individual_methods}
\end{figure}

However, we have found that there are several general-purpose baselines that can 
often outperform proposed speedup methods. In Figure~\ref{fig:individual_methods}, we train ResNet-50 with various speedup methods proposed in the literature and available in Composer \citep{composer}. We chose these methods because all had tested implementations in a common library. Moreover, all but one were included in the winning submission to the 2022 MLPerf Open Division (using these same implementations), suggesting that these speedups have an unusually large degree of empirical support. See Appendix~\ref{app:experiments} for more details.

The simple baseline of training for less time outperforms nearly all proposed methods. Crucially, this is only true when the learning rate schedule is adjusted along with the training time. Simply aborting a single training run early---the ``Early Stopping'' baseline---is far less effective. This discrepancy between a full, short training run and an aborted, long training run is consistent with the findings of \citet{chinchilla}.

Using a smaller model can also be a solid baseline. In this case, we find that training a ResNet-18 or ResNet-34 rather than a ResNet-50 can outperform some speedup methods with certain hyperparameters. For simpler architectures (e.g., most recent NLP architectures), one could also reduce the depth or width algorithmically.

These results do not indicate that the tested speedup methods are necessarily ineffective. This is because changing the hyperparameters, model, dataset, optimizer, etc. could alter the efficacy of these methods and the baselines. E.g., if we were to make the model $1000\times$ larger and train for a single epoch, it is likely that reducing model size would be a more effective speedup than further reducing training time. 

What these results do indicate is that simple approaches like reducing the epoch count and model size can be strong baselines. We, therefore, recommend that authors use these baselines in their speed-accuracy tradeoff curves.

\paragraph{Account for Composition}


Along the same lines as holding hyperparameters and libraries constant when comparing methods, one must also hold the set of speedup methods constant. This is because a given method may be more or less effective depending on what other methods are present.

\begin{figure}[t]
\centering
\includegraphics[width=1\linewidth]{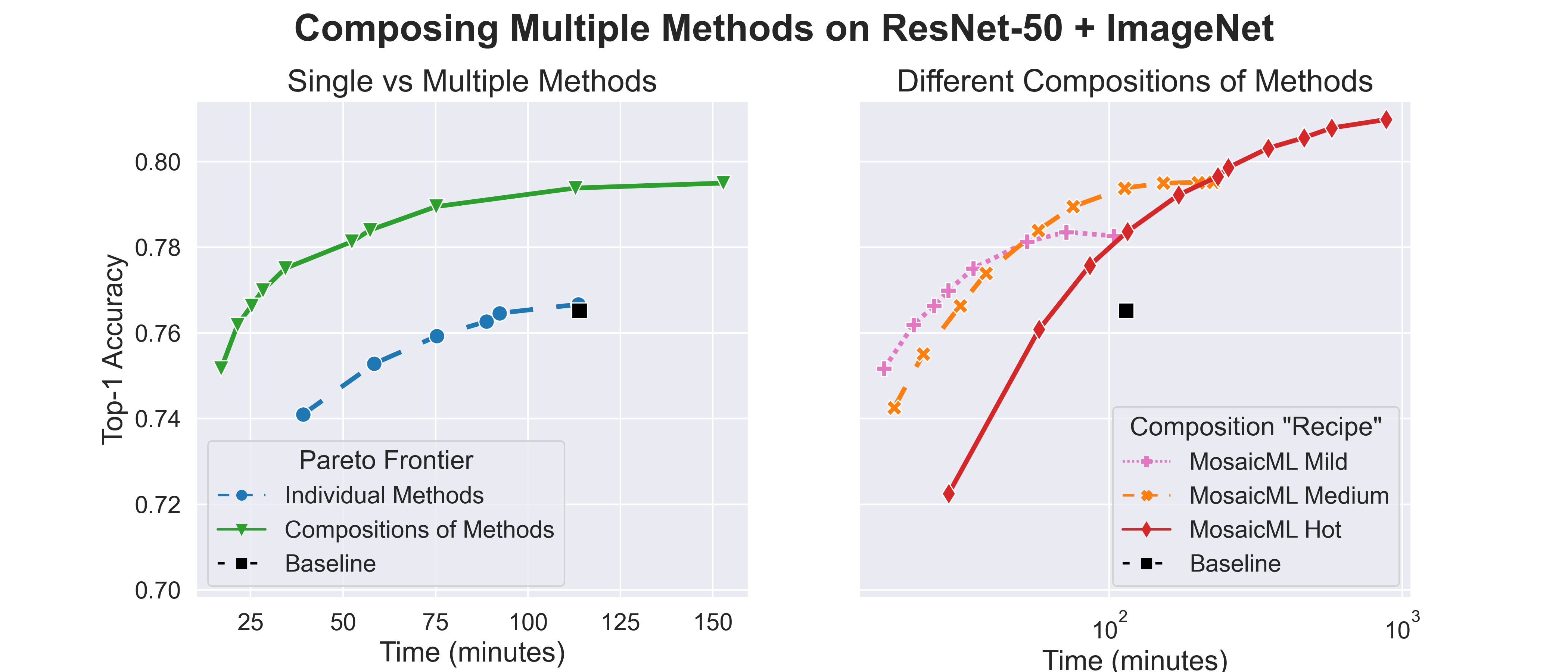}
\caption{\textit{(left)} Composing many methods for speeding up training can work far better than using any individual method. \textit{(right)} Different combinations of methods are effective for different amounts of the training budget. No combination is universally the best.}
\label{fig:composition}
\end{figure}

This is illustrated in Figure~\ref{fig:composition}, where we compare results across the fastest known ResNet-50 training ``recipes'' \citep{mosaic2022rn50} and show the time-accuracy Pareto frontier of the methods from Figure~\ref{fig:individual_methods}. Each recipe obtains a different time vs. accuracy tradeoff, with different points within a curve representing different training runs with varying epoch counts. Details about these recipes are given in Appendix~\ref{app:experiments}.

The most obvious pitfall is failing to account for the benefits of other methods. In Figure~\ref{fig:composition} (left), we see that simply composing multiple methods improves the time vs. accuracy curve greatly compared to the curve made from the best points attainable by using any single method. If a new paper uses more algorithmic speedup methods in its baseline than an older paper, the former might incorrectly attribute its better numbers to its proposed method rather than its algorithmically-enhanced baseline setup.



A subtler source of error is that different methods are effective for different amounts of training time. In Figure~\ref{fig:composition} (right), the ``MosaicML Medium'' and ``MosaicML Hot'' recipes differ only in the presence of regularization methods (see Appendix~\ref{app:experiments} for details). These methods harm accuracy at short training times but help it at long training times.

\paragraph{Ensure Reliability and Significance of Results}

We recommend that authors perform multiple experimental runs with separate initializations and random seeds. When computationally feasible, results should include clearly defined error bars and a measure of central tendency (e.g., mean) and variation (e.g., standard deviation). As an alternative to reusing the same settings multiple times with different seeds, one can also evaluate more points in the speed vs. quality tradeoff curve (as we have done in the previous subsections).

All the comparisons should include a description of how the comparisons were produced (i.e., whether data was taken from a paper, a paper was reimplemented, or a paper's code was used) and any differences between or uncertainties about the baseline setting and the setting used for experiments.


\section{Guidelines for Achieving Speedup in Practice} \label{sec:guide}

In order to speed up deep learning training and correctly interpret one's results, it is important to be aware of the various hardware resources involved in training. Without this knowledge, it is easy to draw incorrect conclusions or accelerate operations that are not actually the bottleneck, which provides no benefit.

\begin{figure}[t]
\centering
\includegraphics[width=0.5\linewidth]{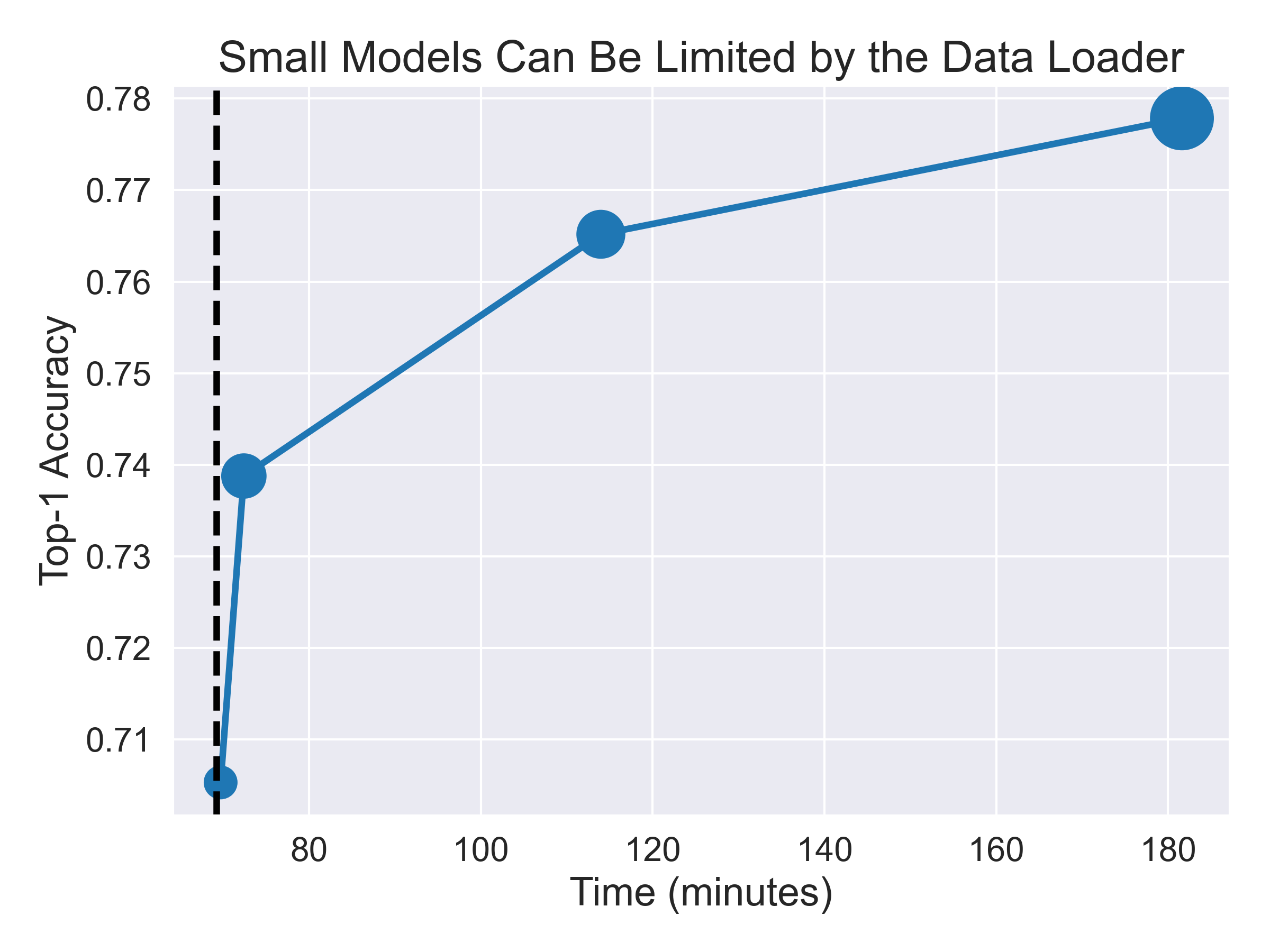}
\caption{Loading data can often be a bottleneck when dealing with images, especially for small models. Shown are training time and accuracy for ResNet-18, ResNet-34, ResNet-50, and ResNet-101. Marker sizes are proportional to parameter counts. The vertical line indicates the fastest time any model could achieve given the data loader throughput. Failing to control for the data loader bottleneck can make small models appear worse and prevent ``speedup'' methods from yielding speedups.}
\label{fig:dataloader}
\end{figure}

As a motivating example, consider Figure~\ref{fig:dataloader}. The curve shows the accuracy and 90-epoch training times for ResNet-\{18,34,50,101\} on ImageNet. At first glance, it appears that there is a large accuracy benefit and little speed cost when moving from ResNet-18 to ResNet-34. This could lead one to conclude that, e.g., ResNet-34 is a better architecture than ResNet-18, or that doubling depth is an excellent design choice. However, the ResNet-18 result is handicapped by the data loader. It \textit{appears} not much faster than ResNet-34, but only because the training speed is insensitive to model size in this regime. If we loaded data more quickly, ResNet-18 could be much faster.\footnote{This example is not contrived. Because these results use FFCV \citep{ffcv} on a modern CPU with carefully tuned data loader parameters, it is likely that most reported image classification results with small models are even more bottlenecked by the data loader than this.}

Data loading is a common bottleneck in training but by no means the only one. Below are various hardware resources that can become training bottlenecks. This list is not exhaustive---e.g., we omit discussion of caches, register files, execution ports, reorder buffers, read queues, and countless other aspects of various hardware elements. Instead, we focus on a few key resources that all efficiency researchers should be able to reason about for common hardware.

\subsection{Possible Bottlenecks and Mitigation}

\subsubsection{GPU compute}
\paragraph{Overview}
The term ``compute'' is often used imprecisely. For our purposes, ``compute'' refers to operations executed on data, as opposed to the movement of data between devices or levels of the cache hierarchy.
Nearly all accelerators have abundant compute capacity. In particular, modern GPUs have much of their area devoted to special hardware for dense matrix multiplications---the most compute-intensive operations in nearly all neural networks.
As we saw in Figure~\ref{fig:speed_proxies}, this makes dense multiply-add operations much cheaper than other computations like normalization, nonlinearities, sparse operations, etc. As a result, compute tends to only be the bottleneck when performing large matrix multiplications or convolutions.

\paragraph{Mitigation}
The simplest way to eliminate this bottleneck is to reduce the sizes of the tensors involved. This can mean a reduction in feature/channel count or a reduction in resolution/sequence length. One can also impose block sparsity with a large block size, 2:4 sparsity on recent NVIDIA GPUs, or low-rank/structured parameter matrices. See Section~\ref{sec:funcSpeedups} for specific approaches.

\subsubsection{GPU memory}
\paragraph{Overview}
A GPU can only store so much data. During training, memory usage is typically dominated by the optimizer state and activations. The optimizer state is usually a constant factor larger than the model itself, and so dominates when the batch size times input size is small. Here ``input size'' refers to the resolution, sequence length, or other appropriate measurements. When the batch size times the input size is large, the activation memory typically dominates. E.g., a float32 ResNet-50 model takes around 100MB to store, but its activations can require many gigabytes.

\paragraph{Mitigation}
The model and optimizer state can be sharded across different accelerators \citep{rajbhandari2020zero} and/or offloaded to CPU RAM \citep{zeroOffload}. One can also use an optimizer with a smaller state \citep{dettmers20218, adafactor}. Using a smaller model can also reduce the size of both the model and optimizer state. To reduce activation memory, one can checkpoint/re-materialize activations (ideally intelligently, see \citeauthor{jain2020checkmate}, \citeyear{jain2020checkmate}), shard saved activations \citep{rajbhandari2020zero}, compress activations \citep{actnn, evans2021ac}, choose activation functions like ReLU that require saving only one bit of state per activation, or use in-place normalization ops \citep{inplaceBatchnorm}, among other possibilities (see Sections~\ref{sec:architecture} and \ref{sec:derived}).

\subsubsection{GPU memory bandwidth}
\paragraph{Overview}
Processors, including GPUs, cannot immediately operate on data stored in memory. Instead, values must first be loaded into registers or some other small, fast storage. This loading process takes time. For large blocks of data, such as neural network activations and weights, the loading time is usually limited by the \textit{memory bandwidth}, expressed in bytes per second. A similar cost must also be paid for storing data (i.e., moving it out of fast storage and back into memory).

\paragraph{Mitigation}
To reduce memory bandwidth consumption, one must reduce the size of the input and output tensors for one's operations. This means compressing parameters, activations, and optimizer states (see Sections ~\ref{sec:architecture} and \ref{sec:derived}).
Alternatively, one can \textit{hide} memory accesses by performing more compute per byte. This means \textit{increasing} the sizes of one's inputs and outputs, especially along whichever axis is currently smallest. E.g., if multiplying a $4096 \times 64$ and a $64 \times 8192$ matrix, increasing the contraction dimension from 64 to 128 would increase the ratio of compute to memory accesses (the \textit{arithmetic intensity}) more than increasing the number of rows or columns. To see this, consider that each element of the left matrix is reused 8192 times and each element of the right matrix is reused 4096 times, while each element of the output matrix has only a 64-element dot product's worth of compute per memory write.

\subsubsection{CPU compute}
\paragraph{Overview}
For many applications, preprocessing and/or augmenting data happens in the CPU. This is especially likely to be a bottleneck in computer vision applications, where JPEG decoding and complex data augmentation pipelines can be expensive. CPUs may also be taxed by coordinating communication between GPUs, CPU memory, networking cards, and other devices.

\paragraph{Mitigation}
If a CPU is bottlenecked by data loading, the surest fixes are to simplify one's data loading pipeline (e.g., reducing the number of augmentations), use a higher-quality data loading library such as \href{https://docs.nvidia.com/deeplearning/dali/user-guide/docs/}{NVIDIA DALI} or FFCV \citep{ffcv}, or offload some of the computation to accelerators. This offloading is possible with DALI and/or custom code in one's training loop. Separately, this bottleneck can be \textit{hidden} by performing more compute on the GPU per batch loaded, such as by increasing the complexity of the architecture or optimizer (see ``retrofit'' techniques in Sections \ref{sec:funcSpeedups} and \ref{sec:optimization}). This can provide a speedup through reduced $n_{\text{iteration}}$.

\subsubsection{CPU memory}
\paragraph{Overview}
CPU memory is most often used to store datasets so that samples can be loaded into the GPUs without needing to access storage devices. Without careful implementation, it can also be a limiting factor when saving, loading, or initializing models---for example, multi-GPU training in PyTorch creates one copy of the model per GPU in the machine by default. CPU memory can also be used to offload model parameters, optimizer state, and saved activations \citep{zeroOffload}.

\paragraph{Mitigation}
To avoid the need to cache the full dataset in RAM, one can ensure that the data is kept in fast storage, preferably a modern NVME solid-state drive. To avoid materializing many copies of a model in memory, one can carefully read the documentation for their deep learning tools (see, e.g., the \href{https://pytorch.org/torchdistx/latest/fake_tensor.html}{meta device} in PyTorch). To avoid running out of memory when offloading tensors to the CPU, consider offloading fewer or smaller tensors---e.g., offload only the optimizer state but not activations. One may also be able to reduce the size of their dataset using methods from Section~\ref{sec:traindata}, or the size of their model using methods from Section~\ref{sec:architecture}.

\subsubsection{CPU memory bandwidth}
\paragraph{Overview}
CPU memory bandwidth is not typically a bottleneck in deep learning training on accelerators. However, if training on a CPU, the considerations for GPU memory bandwidth instead apply to CPU memory bandwidth.
\paragraph{Mitigation}
If training on a CPU, the mitigations for GPU memory bandwidth apply.

\subsubsection{Inter-GPU interconnect}
\paragraph{Overview}
In a given machine, GPUs are typically connected to each other and the CPU(s) through a motherboard’s PCIe slots. The PCIe interface is used when tensors are moved between the CPU(s) and GPU(s) and when GPUs within a machine share gradients with each other. For some GPUs, there are also interfaces like NVLINK that allow faster communication between GPUs.

\paragraph{Mitigation}
To reduce the communication cost of gradient synchronization, one can use any number of gradient compression schemes---though they may not be worthwhile \citep{gradCompressionSucks}. One can also synchronize gradients less frequently \citep{dutta2018slow, ortiz2021trade, stich2018local, lin2018don}, though with a possible accuracy penalty. As an alternative, one can increase one's ratio of compute to parameter count by using larger batch sizes or input sizes. There are also various parallelism schemes \citep{shazeer2017outrageously, deepspeedMoE, deepspeed3dblog, narayanan2021efficient}, each with their own tradeoffs, that may be beneficial. If one is sharding data across GPUs, sharding less data (e.g., only optimizer states) may be beneficial. See Sections~\ref{sec:derived} and \ref{sec:optAlgo} for methods that reduce gradient sizes or communication requirements.


\subsubsection{CPU-GPU interconnect}
\paragraph{Overview}
Supplying data to the GPUs is unlikely to be bottlenecked by PCIe, even for tasks like image classification that often have high ratios of input size to model size. However, repeated transfer of data between CPU and GPU (most often due to poor data augmentation implementations) can cause this to be a bottleneck. This interconnect can also be a bottleneck when offloading weights, activations or optimizer states to the CPU \citep{zeroOffload}.

\paragraph{Mitigation}
One can ensure that their data processing pipeline, debugging statements, and other codes do not move data between the CPU and GPU more than needed. If offloading data to the CPU, one can offload less data (e.g., only optimizer states).

\subsubsection{Inter-node interconnect}
\paragraph{Overview}
When carrying out distributed training, the participating machines must communicate with each other to share information. In the common case of data-parallel training with synchronous gradient-based optimizers, this information consists of gradients for all parameters. Note that it is not necessary to transmit the optimizer state so long as this state is a deterministic function of the gradients. This interconnect can also become a bottleneck when sharding saved activations, model parameters, or optimizer states \citep{rajbhandari2020zero}.

\paragraph{Mitigation}
The mitigations here are similar to those for Inter-GPU interconnect, though there can be differences when using different parallelism schemes inter-node and intra-node. Also, when training on multiple nodes, one has the option of purchasing better network cards and switches---this is not possible within a node.

\subsubsection{Storage capacity}
\paragraph{Overview}
A storage device, such as a solid-state drive (SSD) or hard drive (HDD), can only store so much data. This can become a limitation when attempting to store large datasets or model checkpoints.

\paragraph{Mitigation}
The most straightforward solution is to obtain larger or more storage devices. Alternatively, one can use a data loading library or filesystem that allows streaming in data as needed. One can also attempt to subsample the dataset (see Section~\ref{sec:traindata}). For large model checkpoints, reducing the size of the model (Section~\ref{sec:architecture}) or optimizer state (Section~\ref{sec:optAlgo}) is helpful.

\subsubsection{Storage bandwidth}
\paragraph{Overview}
An SSD or HDD can only supply data to the CPU so fast. If the storage is too slow, it can cause the data loading to be the bottleneck in the training loop.

\paragraph{Mitigation}
One can split data across more local storage devices, preferably in a RAID \citep{raid} configuration that allows parallelizing reads across devices. With a sufficiently fast network connection and remote storage, streaming data in as needed can also be effective. If one is willing to alter the semantics of their training run, Data Echoing \citep{choi2019faster}, using a larger model, or using more expensive optimization---e.g., \citet{foret2020sharpness,du2021efficient,yao2021adahessian,scalableKFAC,pauloski2021kaisa,distributedShampoo}---can help conceal this bottleneck: see Sections~\ref{sec:architecture} and \ref{sec:optimization}.

\subsubsection{Storage network bandwidth}
\paragraph{Overview}
Similarly, even if a storage device itself is fast, reading data from it can be a bottleneck if the device is connected through a slow network connection (e.g., when using distributed filesystems or remote object stores).

\paragraph{Mitigation}
The ideal mitigation is to instead store data locally and/or cache it in local RAM. If that isn't possible, the same techniques for alleviating a storage bandwidth bottleneck also apply.

\subsection{Overview of Throughputs} \label{app:throughput}
In order to reason about the bottlenecks in a given training workload, it is helpful to know roughly how much data each computational resource can move or consume in a given amount of time. We provide the peak throughput of each of the aforementioned resources on representative hardware in Table \ref{tab:resource_thruputs}.

\begin{table}[h]
    \centering
    \begin{tabular}{l|c}
        \hline
        Resource & Number per second (trillions) \\
        \hline
        A100 40GB SXM f16 dense multiply-adds & 156 madds  \\
        A100 40GB SXM f32 scalar instructions & 9.75 FLOPs \\
        Intel Xeon Platinum 8380 CPU f32 dense multiply-adds & 2.9 madds \\
        A100 40GB SXM memory bandwidth & 1.56 bytes \\
        A100 40GB SXM NVLink bandwidth & 0.3 bytes \\
        High-end (800Gbps) inter-node interconnect & 0.1 bytes \\
        PCIe v4 bandwidth & 0.032 bytes \\
        DDR4-3200 RAM memory bandwidth & 0.0256 bytes \\
        Sabrent Rocket 4 Plus SSD read bandwidth & .007 bytes \\
        Low-end (10Gbps) inter-node interconnect & .00125 bytes \\
        \hline
    \end{tabular}
    \caption{Computational elements differ in peak throughput by many orders of magnitude. Unless a workload uses these elements in parallel and in proportion to their throughput, some resources will be underutilized while some will be bottlenecks. It is not safe to assume that multiply-adds are the bottleneck.}
    \label{tab:resource_thruputs}
\end{table}

While the exact numbers will vary based on one's hardware configuration, it is clear that there are extreme differences across computational elements. For example, a single A100 can perform $156 / (.00125 / 2) = 249600$ multiply-adds for every 16-bit value transmitted over a 10 gigabit network connection. Furthermore, when considering entire machines, the differences might be even starker; e.g., with an 8 GPU server, the total multiply-add throughput will increase above 1 quadrillion per second, while machine-level resources like the inter-node interconnect will not increase.

These enormous differences in throughput across devices highlight the importance of incorporating an understanding of hardware into one's application, evaluation, and research of algorithmic speedup methods.

\section{Summary and Path Forward}
\label{sec:conclusion}
We formalized the training speedup problem and devised a taxonomy through which various speedup approaches can be understood.
Using this taxonomy, we surveyed over 200 approaches to accelerating neural network training, classifying methods in terms of fundamental speedup building blocks. Based on observed trends---and experiments with speedup methods and analysis of compute platforms we introduced here---we additionally provided methodological evaluation and practical application guides. Our guidance is enhanced through the connections it makes to the taxonomy and survey, further demonstrating their effectiveness in organizing the literature for the identification of present trends and future opportunities.

Indeed, there are numerous open problems and worthwhile directions for speedup research that we identified. Thus, we conclude with a wide view of the opportunities in this space:

\begin{itemize}
    \item Speedup methods are often devised for and tested on computer vision problems. Transferring these ideas to large language model (LLM) training, generative modeling, reinforcement learning, adversarial training, and many other problems could be valuable for the community.
    
    \item Many subcomponent-action pairs have not been thoroughly explored (see Table~\ref{tab:class}).

    \item Existing approaches usually target one or two subcomponents to achieve a speedup. Approaches that target more could be useful.

    \item Designing new ranking metrics to comprehensively compare speedup methods will aid their evolution. Ranking metrics from the multi-objective community can be the starting place for this research theme. 
    
    \item A good practice for speedup research involving weight initializations is making pretrained models publicly available following FAIR principles (i.e., findability, accessibility, interoperability, and reusability). A more systematic effort toward building a comprehensive Model Zoo will avoid duplication of efforts in many cases. 
    
    \item Some methods have the potential to provide speedups but are hardware-unfriendly, e.g. unstructured sparsity. Algorithm-hardware co-design (i.e., jointly finding an optimal architecture and hardware accelerator) may allow such methods to attain the training efficiency improvements they hint at.
    
    \item A crucial building block for speedup research is a benchmarking leaderboard that can evaluate algorithmic speedup methods in a comprehensive, fair, and reliable manner. Such an effort will propel efficient training advances---e.g., related efforts include DawnBench~\citep{coleman2017dawnbench} and MLCommons Algorithmic Efficiency.\footnote{See \url{https://github.com/mlcommons/algorithmic-efficiency} for  more information.}
    
    \item Finally, our review strongly suggests that cross-pollination of ideas among diverse research communities---e.g., deep learning, blackbox \& multi-objective optimization, approximate computing, and hardware design---is critical to the efficient maturation of a speedup method toolbox.
    
\end{itemize}

\begin{acks}
We thank our reviewers and editor for helpful comments. This work was performed under the auspices of the U.S. Department of Energy by Lawrence Livermore National Laboratory under Contract DE-AC52-07NA27344 and LLNL LDRD Program Project No. 23-ER-030 (LLNL-JRNL-840984).
\end{acks}

\renewcommand{\theHsection}{A\arabic{section}}

\appendix \label{appendix}

\section{Motivation for Compute-Efficient Deep Learning}
\label{app:motivation}
Below we provide a detailed discussion of various limitations that motivate the need for compute-efficient deep learning.  

\paragraph{Computing and Hardware Limitations:}
The doubling of transistor counts every two years predicted by Moore's Law \citep{moore1965cramming} suggests one might expect a corresponding doubling in application performance for roughly the same hardware cost. In the ``Pre Deep Learning Era'' (i.e., before 2010), the amount of computation required to train ML models grew in line with Moore's Law, doubling roughly every 20 months. However, the computational burden of ML training has outpaced Moore's Law since then. 
In the ``Deep Learning Era'' (i.e., 2010 to 2015), the scaling of computing power for model training significantly accelerated to doubling approximately every 6 months. In the ``Large Model Era'' (i.e., after 2015), computing requirements for training massive models really exploded with a 10 to 100-fold increase \citep{thompson2020computational, schwartz2020green, sevilla2022}. Notably, current large-scale models use 600,000 times more computing power than the noteworthy AlexNet model \citep{krizhevsky2012imagenet} from 2012 \citep{powerai}. By extrapolating the compute costs into the future, \citet{thompson2021deep} showed that it will take $10^5 \times$ more computing to get a 5 percent error rate on ImageNet. In other words, our current trajectory suggests that progress will be hindered by the pace of hardware advancement.

\paragraph{Economic Cost:} Given the fact that compute demands for deep learning are growing significantly faster than the improvements in hardware performance, the training process is becoming more time-consuming and increasingly expensive \citep{sharir2020cost, thompson2020computational, powerai}. 
For example, GPT3 required approximately 3,600 petaFLOPS-days to train. At the advertised maximum performance of a Google TPU v3, it would take approximately \$1.65 million to train GPT3. Interestingly, a standard laptop needs about a year to reach one petaFLOPS-day, which implies that it will take several millennia to train GPT-3 \citep{powerai}. Similarly, it was estimated to cost around \$35 million in computing power to replicate the experiments reported in the AlphaGo Zero paper\footnote{\url{https://www.yuzeh.com/data/agz-cost.html}}. By extrapolating the economic costs into the future, \citet{thompson2020computational} showed that continued performance improvements in a range of application domains will take billions to trillions of dollars using current training strategies. Similarly, \citet{powerai} showed that the training cost of the largest AI model in 2026 predicted by their trendline would cost more than the total U.S. GDP, which certainly is infeasible.  

\paragraph{Environmental Cost:} 
Emissions of greenhouse gases such as carbon dioxide or equivalents (CO2eq) due to human activities are the root cause of global warming. Energy production is the major factor in greenhouse gas emissions, contributing around 25\% of emissions in 2010~\citep{tollefson2018ipcc, henderson2020towards}. ML training has the potential to significantly contribute to carbon emissions due to the energy required to power hardware for a long period of time \citep{dodge2022measuring}. \citet{lacoste2019quantifying} and \citet{henderson2020towards} proposed tools to estimate the energy and carbon footprints of ML training.
\citet{strubell2019energy} showed that the training transformer model (with neural architecture search) could produce an estimated 626,155 lbs of carbon dioxide (CO2) emissions, which is equivalent to the lifetime carbon footprint of five cars (126,000 lbs/car). \citet{thompson2021deep} projected that the error level of the best ImageNet classifier would be reduced to just 5\% by 2025 but would require emission of as much carbon dioxide as New York City generates in one month. The authors in \citet{patterson2021carbon, patterson2022carbon} improved the estimates of energy consumption and CO2 emission in the prior studies. Unfortunately, this revised carbon footprint of ML training is still too large to ignore.

\paragraph{Limited Applications:} 
The compute-intensive nature of deep neural network training severely limits the potential application areas DL can make an impact on. Several application domains where highly accurate predictive models can enable unprecedented success use of resource-limited devices. For example, DL-enhanced ocean monitoring could provide scientists with a powerful tool to enhance the pace of oceanographic research \citep{ahmad2019machine}. Similarly, space exploration using autonomous planetary exploration devices endowed with predictive capabilities offers a huge potential to make unprecedented scientific advances \citep{diffenderfer2021winning}. Such resource-limited applications are limitless and can be categorized under the umbrella term Edge-AI \citep{murshed2021machine}. Existing Edge-AI relies upon a centralized server for training and updating the ML model, which results in low coverage/accuracy, limited adaptability, and high latency, especially in applications with high spatiotemporal variations. While doing efficient inference at the edge is quite common, training at the edge is still not feasible due to the compute-intensive nature of current training methodologies \citep{bhardwaj2022benchmarking, bhardwaj2022unsupervised}. These limitations are holding us back from exploiting our advanced sensing capabilities to their full potential and, in turn, achieving game-changing societal advances.

\paragraph{Democratization Obstacles:}
The democratization of AI aims to make it possible to create and use deep learning for everyone. There is a growing consensus among researchers and policymakers to enable techniques so that the benefits of this technology are not limited to a small group of people \citep{ahmed2020democratization}. Unfortunately, deep learning research is increasingly becoming more computationally intensive, which favors a few resource-rich organizations. This concentration of power can lead to marginalization, causing severe inequalities \citep{ahmed2020framework}.
\citet{ahmed2020democratization} analyzed 171,394 papers from 57 prestigious computer science conferences and concluded that there is a divergence between the following two groups--large firms and non-elite universities. This rise in divergence is driven by access to computing power, which they termed the ``compute divide''. The compute-intensive nature of deep learning training presents an obstacle to ``democratizing'' AI and increases concerns around bias, fairness, and marginalization.

\section{Experimental Details}
\label{app:experiments}
All models were trained on a single machine with eight A100s and two 32-core AMD EPYC 7513 processors. The only models that were not ResNet-50 instances were those corresponding to the ``Smaller ResNets'' baseline in Figure~\ref{fig:individual_methods}.

\begin{figure}[h]
\centering
\includegraphics[width=0.5\linewidth]{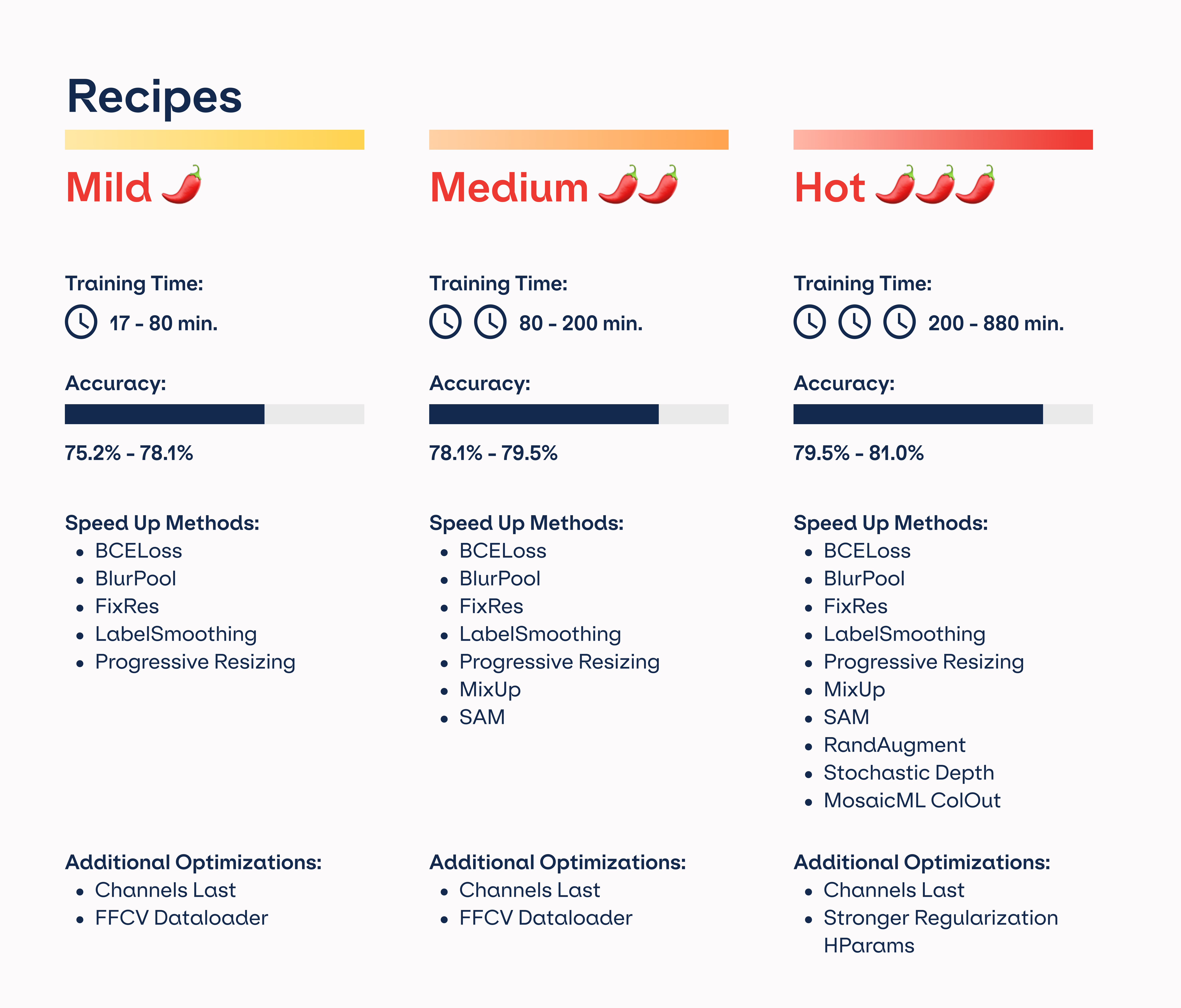}
\caption{Speedup methods used for different curves in Figure~\ref{fig:composition}. See \cite{mosaic2021method} for details about each method.}
\label{fig:mosaic-resnet-recipes}
\end{figure}

The ``Mosaic ResNet'' results were taken from the authors' \href{https://app.mosaicml.com/imagenet?sortBy=costSameQuality&model=resnet50&cloud=mosaicml&hardware=a100_80gb&algorithms=all&baseline=r50_optimized_p4d&recipe=mosaicml_baseline&recipe=mosaicml_hot&recipe=mosaicml_medium&recipe=mosaicml_mild&recipe=ffcv}{public data}. We chose the hyperparameters for speedup methods in Figure~\ref{fig:individual_methods} based on \href{https://github.com/mosaicml/benchmarks/tree/4f1f093c822ffdc18c5dc62206ed235681b0043b/blogs/resnet/recipes}{the hyperparameters} used for these recipes. When the hyperparameters did not vary across recipes for a given speedup, we made up similar hyperparameters on a best-effort basis that would allow for assessing alternate speed vs accuracy tradeoffs---e.g., choosing different degrees of progressive resizing. These hyperparameters may not be optimal, so it is important to conclude only that certain baselines \textit{can} outperform these methods, not that they always will. More information about the different sets of methods used for each curve in Figure~\ref{fig:composition} is shown in Figure~\ref{fig:mosaic-resnet-recipes}.

Microbenchmarking experiments used a single A100, with means and standard deviations computed from five trials. Each trial included 10 executions of the operation in question. All times within a trial were measured after running four ``warmup'' executions to avoid one-time CUDA and PyTorch overheads. We time only the forward pass for simplicity. All results use half-precision weights and activations.

\section{Additional Experiments}
\label{app:new_experiments}

\begin{figure}[h]
\centering
\includegraphics[width=0.9\linewidth]{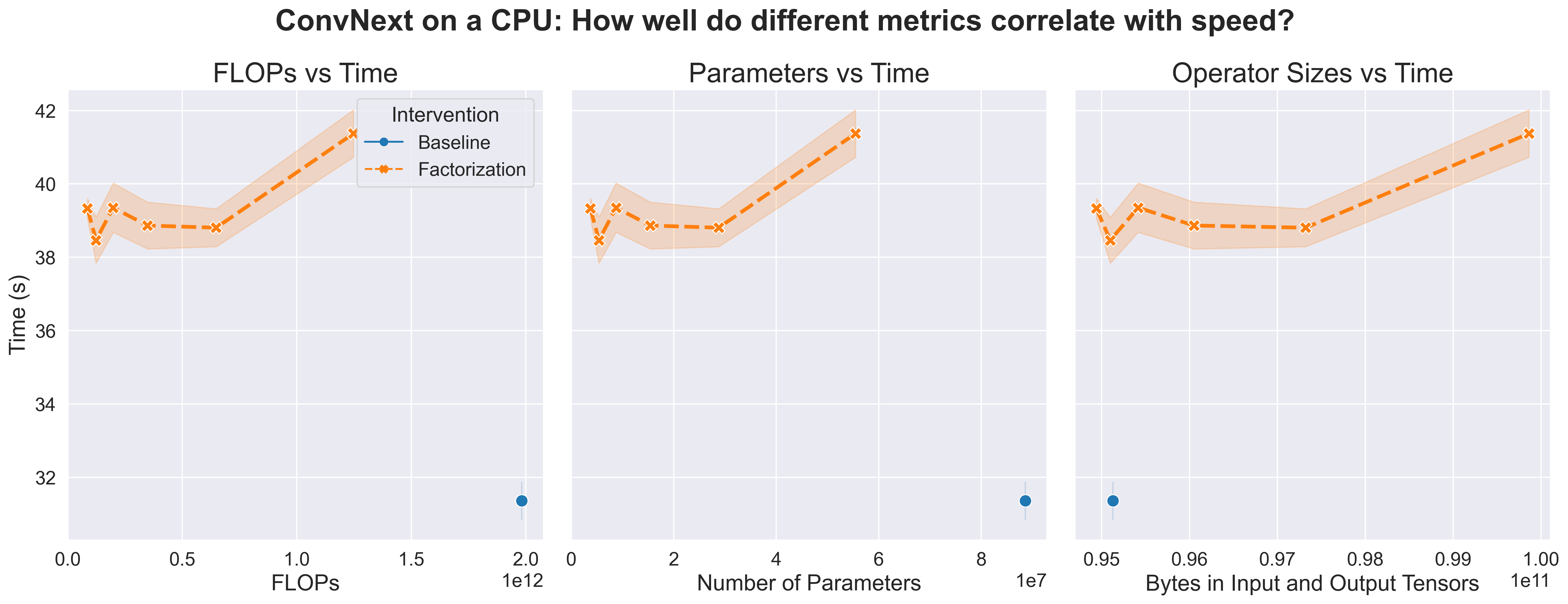}
\caption{Runtime vs various metrics for ConvNeXt-B on 64 CPU cores.}
\label{fig:convnext-cpu}
\end{figure}

\begin{figure}[h]
\centering
\includegraphics[width=0.9\linewidth]{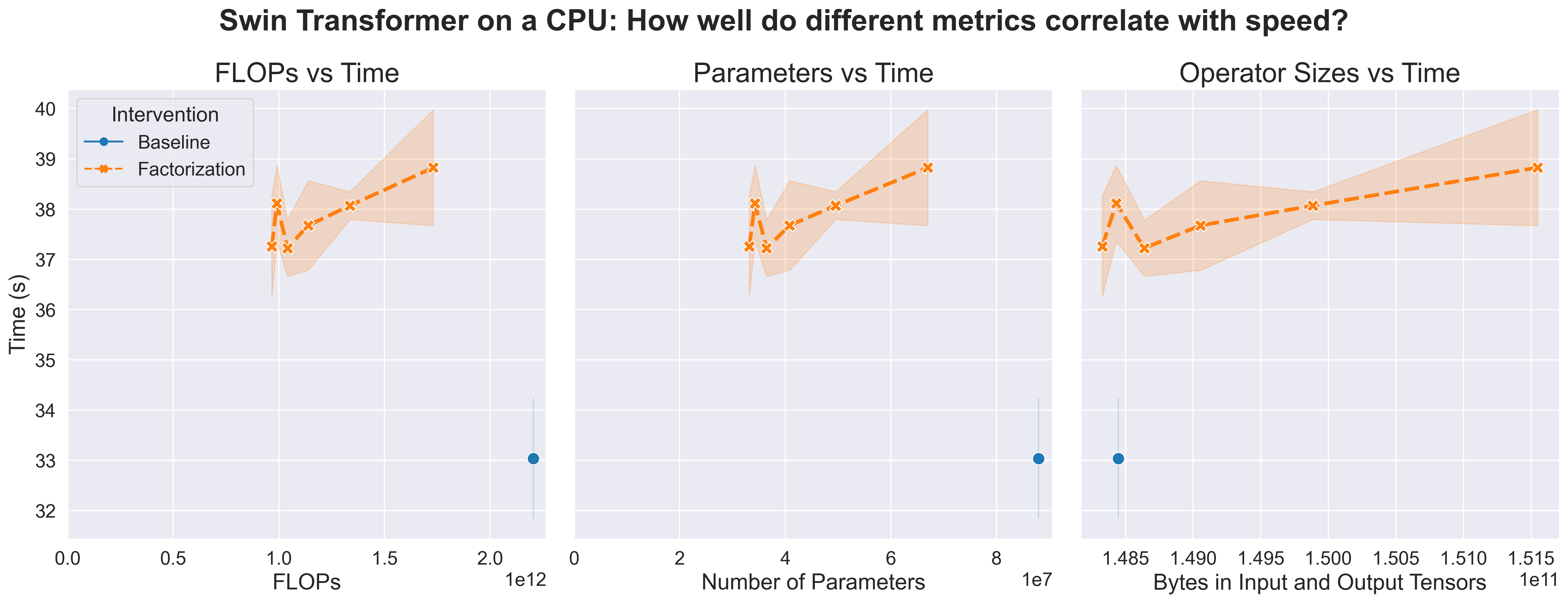}
\caption{Runtime vs various metrics for Swin-B on 64 CPU cores.}
\label{fig:swin-cpu}
\end{figure}

Here, we provide microbenchmarking results for ConvNeXt-B \citep{convnext} and Swin-B \citep{swin} on the CPU to complement our GPU results for these models, shown in Section \ref{sec:pitfalls}. As we found in our experiments on the GPU, the CPU results in Figures \ref{fig:convnext-cpu} and \ref{fig:swin-cpu} show that FLOP count can be a poor proxy for wall time, and memory bandwidth consumption (while imperfect) can be a more helpful proxy.

\bibliography{speedup}

\end{document}